\documentclass[10pt,twocolumn,letterpaper]{article}

\usepackage{cvpr}              % To produce the CAMERA-READY version
\usepackage{graphicx}
\usepackage{listings}
\usepackage{enumerate}
\usepackage{pbox}
\usepackage{url}
\usepackage{amsmath}
\usepackage{amssymb}

\usepackage{algorithm}% http://ctan.org/pkg/algorithm
\usepackage{algpseudocode}% http://ctan.org/pkg/algorithmicx

\usepackage{epsfig}
\usepackage{graphicx}
\usepackage{subcaption}
\usepackage{caption}
\usepackage{makecell}
\usepackage{booktabs}
\usepackage{pifont}
\newcommand{\cmark}{\ding{51}}%
\newcommand{\xmark}{\ding{55}}%

\usepackage[table,x11names]{xcolor}
\usepackage{multirow}
\usepackage{setspace}

\usepackage[misc]{ifsym}

\newcommand{\comment}[1]{}

\newcommand{\hloss}[1]{{\cal L}_{\rm Hungarian}(#1)}

\newcommand{\lmatch}[1]{{\cal L}_{\rm match}(#1)}
\renewcommand{\Sigma}{\mathfrak{S}}

\definecolor{LightCyan}{rgb}{0.88,1,1}

\newlength\savewidth\newcommand\shline{\noalign{\global\savewidth\arrayrulewidth
  \global\arrayrulewidth 1pt}\hline\noalign{\global\arrayrulewidth\savewidth}}
\newcommand\paperurl[1]{{\footnotesize{\color{blue}{\url{#1}}}}}

\newcommand{\noobject}{\varnothing}

\makeatletter
\def\@fnsymbol#1{\ensuremath{\ifcase#1\or \dagger\or \ddagger\or
\mathsection\or \mathparagraph\or \|\or **\or \dagger\dagger
\or \ddagger\ddagger \else\@ctrerr\fi}}
\makeatother

\usepackage{algorithm}
\usepackage[]{algpseudocode}

\usepackage{booktabs}

\usepackage[pagebackref=true,breaklinks=true,colorlinks=true,bookmarks=false]{hyperref}

\usepackage[capitalize]{cleveref}
\crefname{section}{Sec.}{Secs.}
\Crefname{section}{Section}{Sections}
\Crefname{table}{Table}{Tables}
\crefname{table}{Tab.}{Tabs.}

\def\confName{CVPR}
\def\confYear{2023}

\begin{document}

\title{DETRs with Hybrid Matching}

\author{
Ding Jia$^{1}{\thanks{Core contribution.}}$\quad\quad\quad\;
Yuhui Yuan$^{4\dagger}{
\thanks{Corresponding author}}$ \quad\quad\quad
Haodi He$^{2\dagger}$ \quad\quad\quad
Xiaopei Wu$^3$\\
Haojun Yu$^1$ \quad\:
Weihong Lin$^4$\quad\:
Lei Sun$^4$ \quad\:
Chao Zhang$^1$ \quad\!
Han Hu$^{4}$ \\[2mm]
$^1$Peking University \quad\quad
$^2$Stanford University \quad\quad
$^3$Zhejiang University \quad\\
$^4$Microsoft Research Asia
}

\maketitle

\begin{abstract}
One-to-one set matching is a key design for DETR to establish its end-to-end capability, so that object detection does not require a hand-crafted NMS (non-maximum suppression) to remove duplicate detections. This end-to-end signature is important for the versatility of DETR, and it has been generalized to broader vision tasks.
However, we note that there are few queries assigned as positive samples and the one-to-one set matching significantly reduces the training efficacy of positive samples. We propose a simple yet effective method based on a hybrid matching scheme that combines the original one-to-one matching branch with an auxiliary one-to-many matching branch during training. Our hybrid strategy has been shown to significantly improve accuracy. In inference, only the original one-to-one match branch is used, thus maintaining the end-to-end merit and the same inference efficiency of DETR. The method is named $\mathcal{H}$-DETR, and it shows that a wide range of representative DETR methods can be consistently improved across a wide range of visual tasks, including Deformable-DETR, PETRv2, PETR, and TransTrack, among others.
Code is available at:
{\small\url{{https://github.com/HDETR}}}.
\end{abstract}

%%%%%%%%% BODY TEXT

\section{Introduction}
Since the success of pioneering work DEtection TRansformer (DETR)~\cite{carion2020end} on object detection tasks,
DETR-based approaches have achieved significant progress on
various fundamental vision recognition tasks such as object detection~\cite{zhang2022dino,zhu2020deformable,meng2021CondDETR,roh2021sparse}, instance segmentation~\cite{li2022mask,FangQueryInst,dong2021solq,yu2021soit}, panoptic segmentation~\cite{cheng2021masked,li2022panoptic,wang2021max,Yu_2022_CVPR,yuan2020object}, referring expression segmentation~\cite{wu2022language,yang2022lavt}, video instance segmentation~\cite{cheng2021mask2former,wu2021seqformer,wang2021end}, pose estimation~\cite{stoffl2021end,li2021pose,shi2022end}, multi-object tracking~\cite{chen2021transformer,meinhardt2022trackformer,sun2020transtrack}, monocular depth estimation~\cite{li2022depthformer,griffin2021depth}, text detection \& layout analysis~\cite{raisi2021transformer,raisi2022arbitrary,long2022towards,zhang2022text}, line segment detection~\cite{xu2021line}, 3D object detection based on point clouds or multi-view images~\cite{misra2021end,bai2022transfusion,wang2022detr3d,li2022bevformer}, visual question answering~\cite{kamath2021mdetr,lou2022lite}, and so on.

\begin{figure*}[t]
\begin{minipage}{1\linewidth}
\centering
\begin{subfigure}{0.16\textwidth}
{\includegraphics[width=28mm,height=24.5mm]{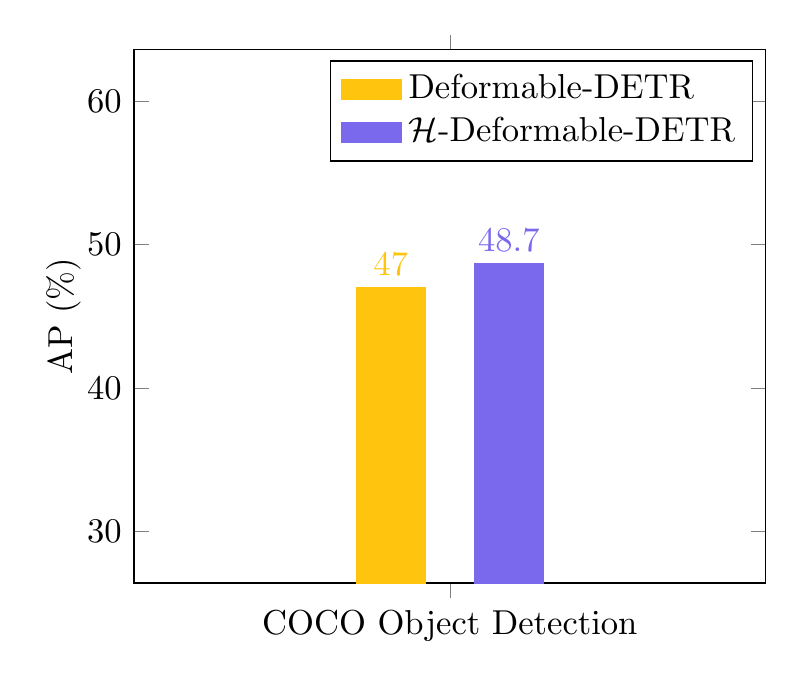}}
\end{subfigure}
\begin{subfigure}{0.16\textwidth}
{\includegraphics[width=28mm,height=24.5mm]{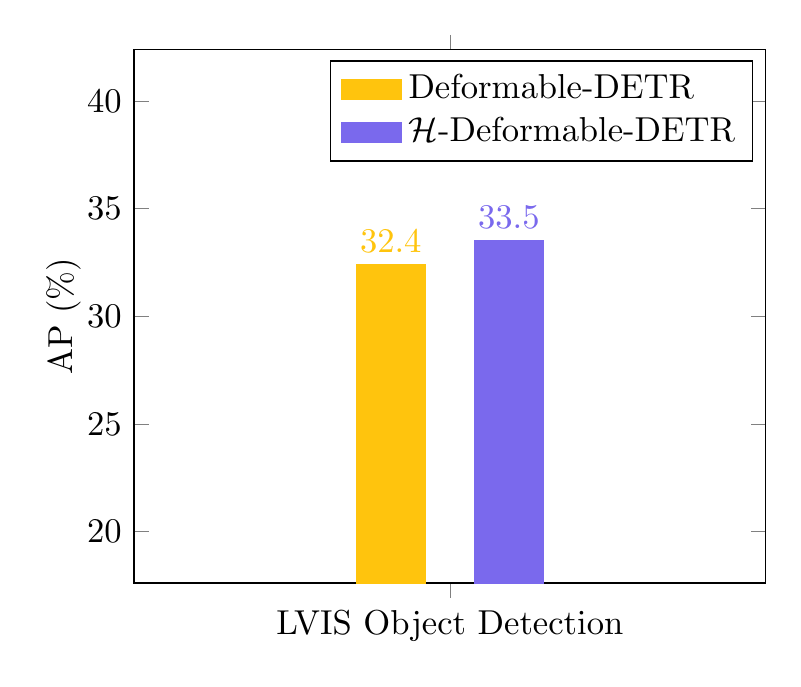}}
\end{subfigure}
\begin{subfigure}{0.16\textwidth}
{\includegraphics[width=28mm,height=24.5mm]{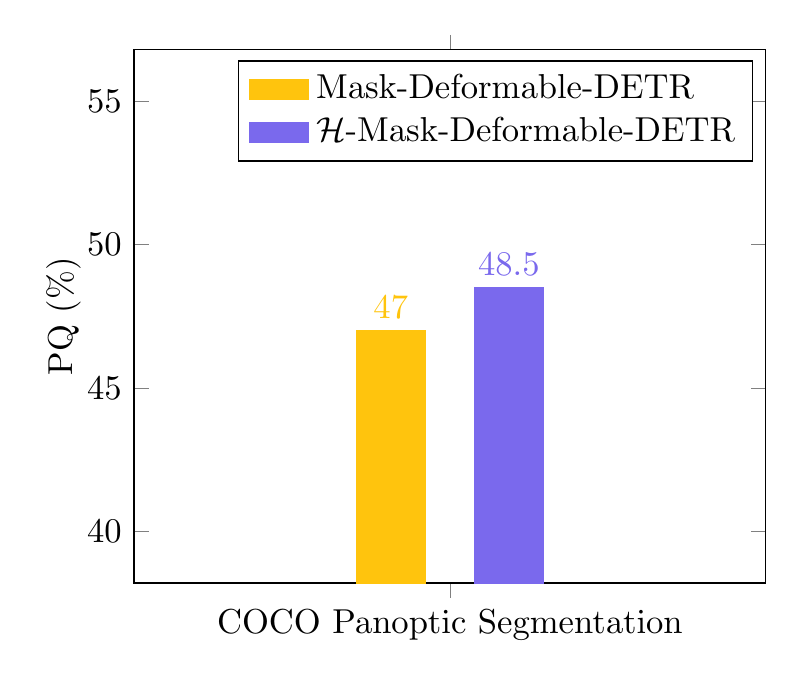}}
\end{subfigure}
\begin{subfigure}{0.16\textwidth}
{\includegraphics[width=28mm,height=24.5mm]{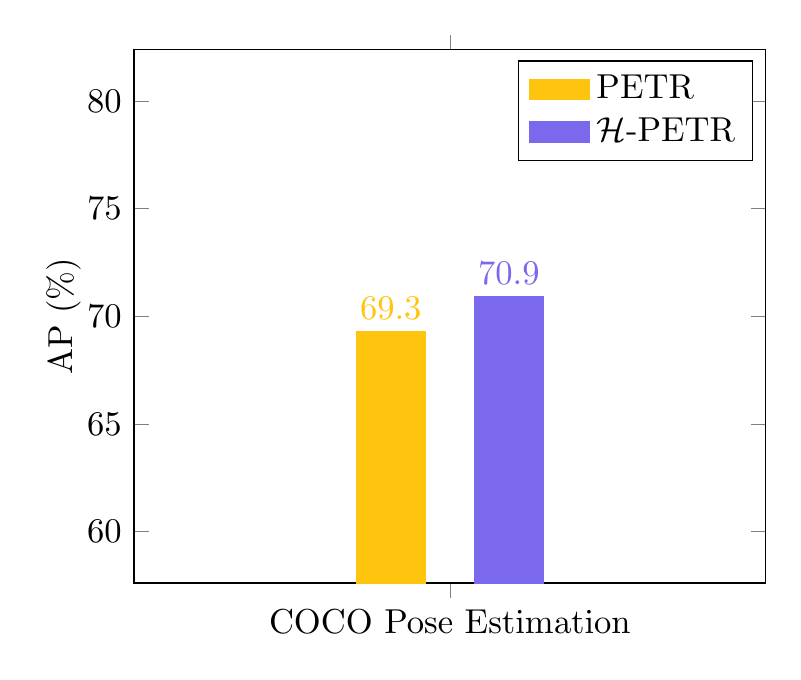}}
\end{subfigure}
\begin{subfigure}{0.16\textwidth}
{\includegraphics[width=28mm,height=24.5mm]{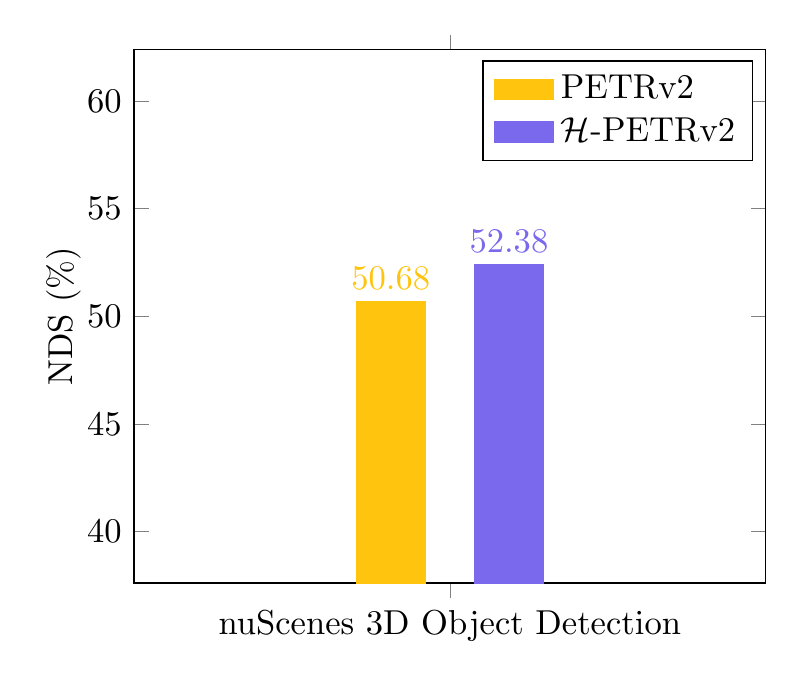}}
\end{subfigure}
\begin{subfigure}{0.16\textwidth}
{\includegraphics[width=28mm,height=24.5mm]{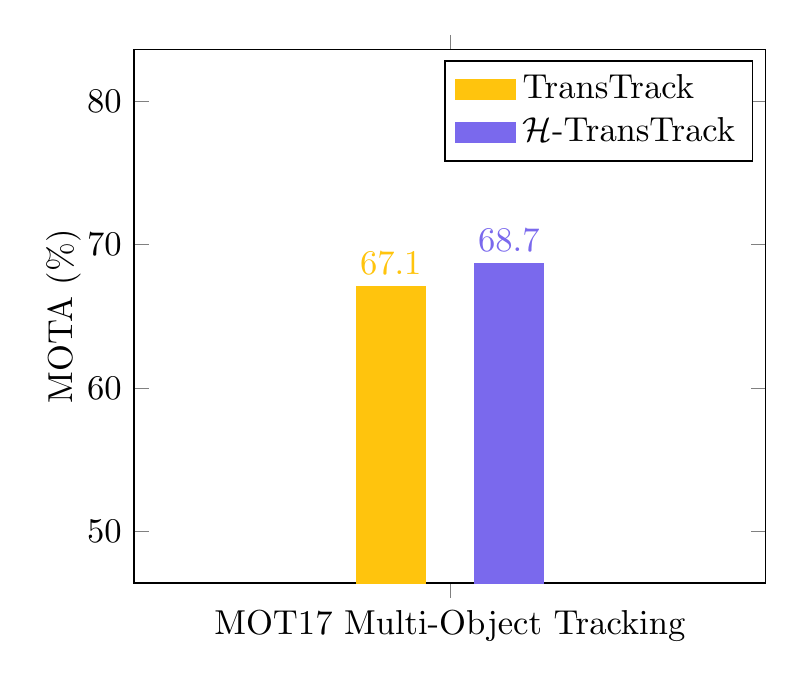}}
\end{subfigure}
\caption{\small{Illustrating the improvements of our hybrid matching scheme across five challenging vision tasks including $2$D object detection, $2$D panoptic segmentation, $2$D pose estimation, $3$D object detection, and multi-object tracking (from left to right). Our hybrid matching scheme gains +$1.7\%$, +$1.1\%$, +$1.5\%$, +$1.6\%$, +$1.7\%$, and +$1.6\%$ over various DETR-based approaches on $6\times$ benchmarks respectively. All the improvements are obtained under the same training epochs and do not require any additional computation cost during evaluation. We choose VoVNetV2~\cite{lee2020centermask}/ResNet$50$ for PETRv$2$/all other methods as the backbone following their original settings.}}
\label{fig:intro_results}
\end{minipage}
\vspace{-2mm}
\end{figure*}

Many follow-up efforts have improved DETR from various aspects, including redesigning more advanced transformer encoder~\cite{zhu2020deformable,gao2021fast} or transformer decoder architectures~\cite{meng2021CondDETR,zhang2022SAMDETR,zhu2020deformable,cfdetr2022,adamixer22cvpr} or query formulations~\cite{wang2021anchor,liu2022dab,li2022dn,zhang2022dino}.
Different from most of these previous efforts,
we focus on the training efficacy issues caused by one-to-one matching, which only assigns one query to each ground truth.
For example, Deformable-DETR typically only selects less than $30$ queries from a pool of $300$ queries to match with the ground truth for each image, as nearly $99\%$ of the COCO images consist of less than $30$ bounding boxes annotations, while
the remaining more than $270$ queries will be assigned as $\noobject$ and are supervised with only classification loss, thus suffering from very limited localization supervision.

{To overcome the drawbacks of one-to-one matching and unleash the benefits of exploring more positive queries,
we present a very simple yet effective hybrid matching scheme, which introduces an additional one-to-many matching branch that assigns multiple queries to each positive sample. 
In inference, we only use the original one-to-one decoder branch supervised with the one-to-one matching loss. We find that this simple approach can substantially improve the training efficacy, especially regarding the fitting of positive queries. Since only the original one-to-one matching branch is used in inference, the merits of the original DETR framework are almost all maintained, for example, avoiding NMS. Our approach also has no additional computation overhead compared to the original version.}

{
We dub the hybrid matching approach as $\mathcal{H}$-DETR, and extensively verify its effectiveness using a variety of vision tasks that adopt DETR methods or the variants, as well as different model sizes ranging from ResNet-50/Swin-T to Swin-L. The visual tasks and the corresponding DETR-based approaches include Deformable-DETR~\cite{zhu2020deformable} for image object detection, PETRv$2$~\cite{liu2022petrv2} for $3$D object detection from multi-view images, PETR~\cite{shi2022end} for multi-person pose estimation, and TransTrack~\cite{sun2020transtrack} for multi-object tracking. The $\mathcal{H}$-DETR achieves consistent gains over all of them, as shown in Figure~\ref{fig:intro_results}. Specifically, our approach can improve the Deformable DETR framework (R50) on COCO object detection by +$1.7\%$ mAP ($48.7\%$ v.s. $47.0\%$), the PETR framework (R50) on COCO pose estimation by +$1.6\%$ mAP ($70.9\%$ v.s. $69.3\%$). In particular, we achieve $59.4\%$ mAP on COCO object detection, which is the highest accuracy on COCO object detection among DETR-based methods that use the Swin-L model. We achieve $52.38\%$ on nuScenes \texttt{val}, which is $+1.7\%$ higher than a very recent state-of-the-art approach of PETRv$2$.
}

\section{Related work}

\vspace{-5mm}
\noindent\paragraph{DETR for object detection.}
With the pioneering work DETR~\cite{carion2020end} introducing transformers~\cite{vaswani17}
to $2$D object detection, more and more follow-up works~\cite{meng2021CondDETR,gao2021fast,dai2021dynamic,wang2021anchor} have built various advanced extensions based on DETR because it removes the need for many hand-designed components
like non-maximum suppression~\cite{neubeck2006efficient}
or initial anchor boxes generation~\cite{girshick2015fast,ren2015faster,lin2017focal,liu2016ssd}.
Deformable-DETR~\cite{zhu2020deformable} introduced the multi-scale deformable self/cross-attention scheme, which attends to only a small set of key sampling points around a reference and achieves better performance than DETR (especially on small objects).
DAB-DETR~\cite{liu2022dab} further verified that a different novel query formulation can also improve the performance.
The follow-up DINO-DETR~\cite{li2022dn,zhang2022dino} has established the new SOTA results on object detection tasks and demonstrated the advantages of DETR design by introducing a novel query denoising scheme.
Different from these works,
we focus on the matching mechanism design of DETR and propose a very simple strategy to improve the training efficacy while still avoiding NMS,
which also differentiates our efforts from the very recent DE-DETR~\cite{wang2022towards} that requires NMS.

\vspace{-3mm}
\noindent\paragraph{DETR for other vision tasks.}
Inspired by the great success of DETR in object detection,
many recent efforts have constructed various DETR-based approaches
for segmentation tasks~\cite{cheng2021masked,cheng2021mask2former,li2022panoptic,yu2022cmt,yu2022k,li2022mask} that aim for more accurate pixel localization and recognition, $3$D object detection tasks~\cite{misra2021end,huang2021bevdet,huang2022bevdet4d,bai2022transfusion,liu2022bevfusion,liang2022bevfusion,liu2022petr,liu2021group,wang2022detr3d,li2022bevformer,liu2022petr,liu2022petrv2} based on point-cloud or multi-view-images that target to identify and localize objects in $3$D scenes, pose estimation tasks~\cite{stoffl2021end,mao2021tfpose,braso2021center,li2021pose,zhang2021direct,li2021tokenpose,shi2022end} with the objective of localizing the key points of the presented persons in a given image, object tracking tasks~\cite{yan2021learning,zhao2021trtr,zeng2021motr,sun2020transtrack,meinhardt2022trackformer} that aim at locating the objects across both spatial and temporal positions within a given video without any prior knowledge about the appearance, and so on.
To verify the generalization ability of our approach,
we first construct a baseline approach, i.e., Mask-Deformable-DETR, for segmentation tasks, and then combine our hybrid matching
scheme with Mask-Deformable-DETR to deliver strong panoptic segmentation results on the COCO benchmark.
For $3$D object detection, pose estimation, and object tracking,
we directly choose the recent approaches including PETRv$2$~\cite{liu2022petr,liu2022petrv2}, PETR~\cite{shi2022end}, TransTrack~\cite{sun2020transtrack} as our baseline
and verifies that our hybrid matching consistently improves their performance.

\vspace{-4mm}
\noindent\paragraph{Label assignment.}
We can categorize the existing label assignment approaches, following the previous work~\cite{wang2021end,zhu2020autoassign}, into two different paths:
(i) \emph{one-to-many label assignment}, i.e., assigning multiple predictions as positive samples for each ground-truth box~\cite{ren2015faster,zhang2020bridging,tian2019fcos}, and
(ii) \emph{one-to-one label assignment}, i.e., assigning only one prediction as a positive sample for each ground-truth box.
POTO~\cite{wang2021end} propose to assign the anchor with either the maximum IoU or closest to the object center as the positive sample, which is modified from the strategies of RetinaNet~\cite{lin2017focal} or FCOS~\cite{tian2019fcos}.
DETR~\cite{carion2020end} and its followups~\cite{meng2021CondDETR,cfdetr2022,wang2021anchor,zhu2020deformable,liu2022dab,li2022dn} apply the Hungarian matching to compute
one-to-one positive assignments based on the global minimum matching cost values between all predictions and the ground-truth boxes.
Different from the most related work POTO~\cite{wang2021end} that only uses one-to-many assignment, based on ATSS~\cite{zhang2020bridging}, to help the classification branch,
our approach chooses Hungarian matching to perform both one-to-one matching and one-to-many matching following DETR and generalizes to various DETR-based approaches across vision tasks.

\vspace{1mm}
\noindent\emph{Relationship to DN-DETR and DINO-DETR:}
{{Our approach is also related to recent approaches that introduce noisy augmentations of ground-truth objects as auxiliary queries, i.e., DN-DETR~\cite{li2022dn} and DINO-DETR~\cite{zhang2022dino}. Similar to our approach, they also introduce additional auxiliary queries. However, the aims of these approaches and ours are different: while DN-DETR and DINO-DETR mainly aim to address the instability issue of Hungarian assignment, we mainly address the insufficient training problem of positive samples in one-to-one matching. }

{
The different aims may have led to their different designs: DN-DETR and DINO-DETR involve a noising scheme and take manual assignment between the noisy queries and ground-truth objects, while our approach is much simpler which uses an end-to-end assignment strategy to match the auxiliary query set the and ground-truth set using the Hungarian method (each ground truth is repeated multiple times).
}
{
The end-to-end manner also allows our approach to be more general than the methods based on denoising query: while DN-DETR/DINO-DETR needs to tune or redesign its noising scheme and query forms, our approach can be easily extended various DETR variants for different vision problems with almost no additional tuning.
}
}

\section{Our Approach}

\vspace{1mm}
\subsection{Preliminary}

\begin{figure}[t]
    \centering
    \includegraphics[width=0.475\textwidth]{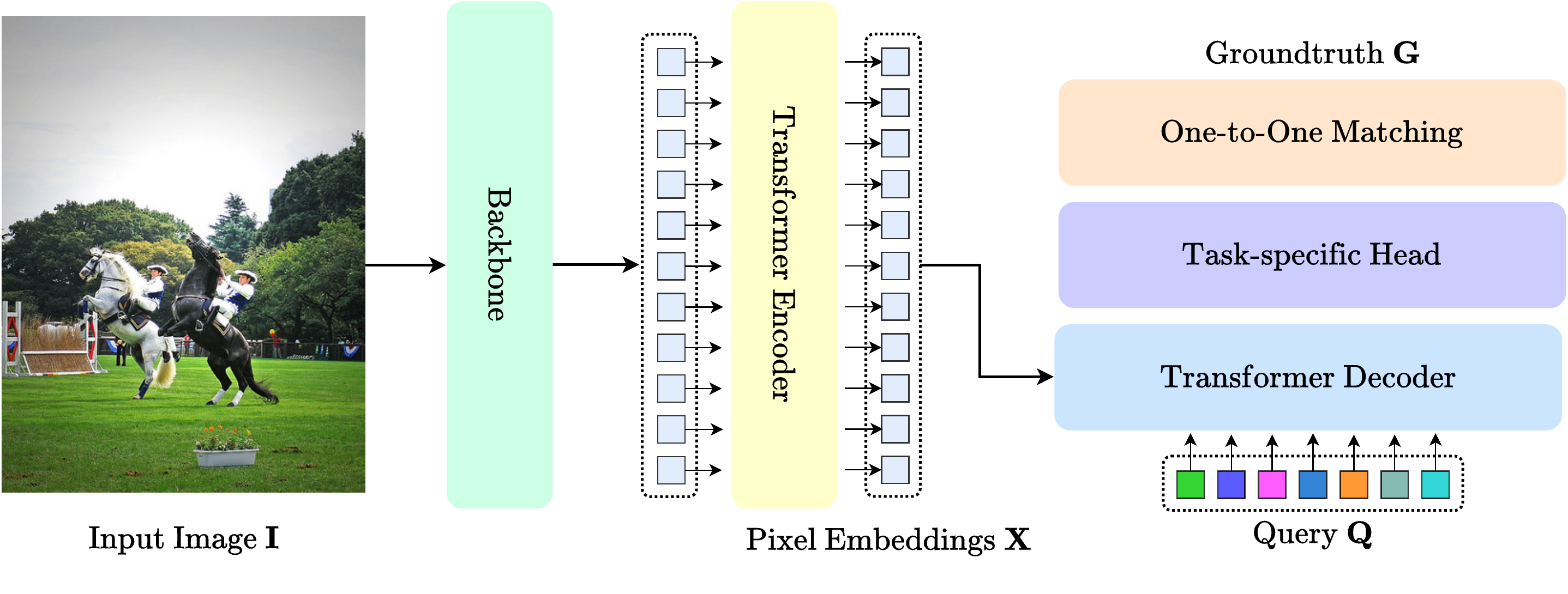}
    \caption{\small{\textbf{Illustrating the pipeline of DETR.}}}
    \label{fig:detr_pipeline}
\end{figure}

\vspace{1mm}
\noindent\textbf{General DETR pipeline.}
Given an input image $\mathbf{I}$, DETR first applies the backbone and the transformer encoder to extract a sequence of enhanced pixel embeddings $\mathbf{X}=\{\mathbf{x}_0,\mathbf{x}_1,\cdots,\mathbf{x}_\textsf{N}\}$.
Second, DETR sends the enhanced pixel embeddings and a default group of object query embeddings $\mathbf{Q}=\{\mathbf{q}_0,\mathbf{q}_1,\cdots,\mathbf{q}_n\}$ into the transformer decoder.
Third, DETR applies the task-specific prediction heads on the updated object query embeddings after each transformer decoder layer to generate a set of predictions $\mathbf{P}=\{\mathbf{p}_0, \mathbf{p}_1,\cdots, \mathbf{p}_n\}$ independently.
Last, DETR performs one-to-one bipartite matching between the predictions and the ground-truth bounding boxes and labels $\mathbf{G}=\{\mathbf{g}_0, \mathbf{g}_1, \cdots,\mathbf{g}_m\}$.
Specifically, DETR associates each ground truth with the prediction that has the minimal matching cost
and apply the corresponding supervision accordingly.
Figure~\ref{fig:detr_pipeline} also illustrates the overall pipeline of the DETR approach.
The follow-up works have reformulated the object query to various variants for different visual recognition tasks such as mask query~\cite{cheng2021masked,li2022panoptic}, pose query~\cite{shi2022end}, track query~\cite{sun2020transtrack,meinhardt2022trackformer}, bins query~\cite{li2022binsformer}, and so on.
We use ``query'' in the following discussions for simplicity and consistency.

\vspace{1mm}
\noindent\textbf{General Deformable-DETR pipeline.}
The Deformable-DETR improves the pipeline of DETR from the following three main aspects: (i) replace the original multi-head self-attention or cross-attention with a multi-scale deformable self-attention and multi-scale deformable cross-attention scheme; (ii) replace the original independent layer-wise prediction scheme with iterative refinement prediction scheme; (iii) replace the original image content irrelevantly
query with a dynamic query generated by the output from the transformer encoder.
Besides, Deformable-DETR also performs one-to-one bipartite matching following the DETR.
Readers could refer to~\cite{zhu2020deformable} for more details.

\subsection{Hybrid Matching}

\begin{figure*}[t]
\centering
\begin{minipage}[t]{0.4\textwidth}
\begin{center}
\includegraphics[width=0.99\textwidth]{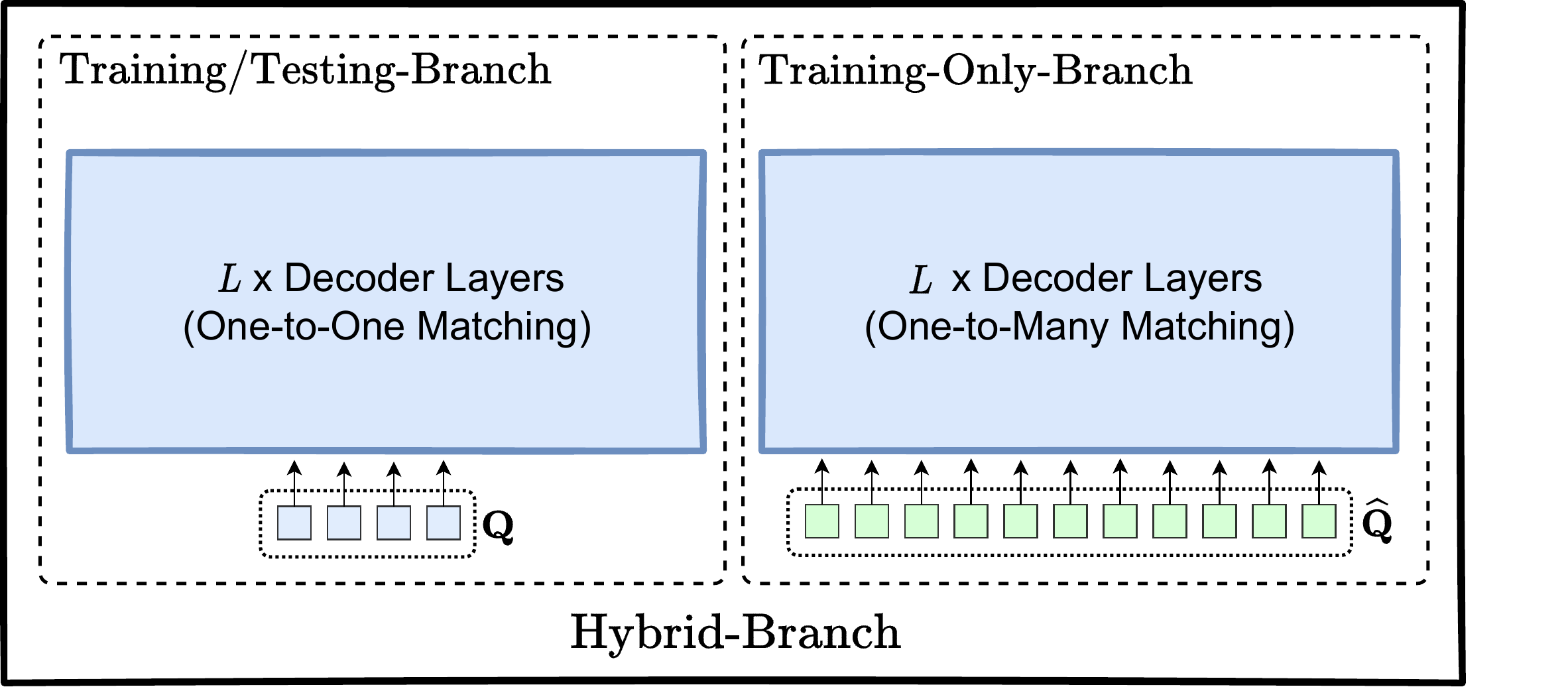}
\end{center}
\end{minipage}
\hspace{-4mm}
\begin{minipage}[t]{0.4\textwidth}
\begin{center}
\includegraphics[width=0.99\textwidth]{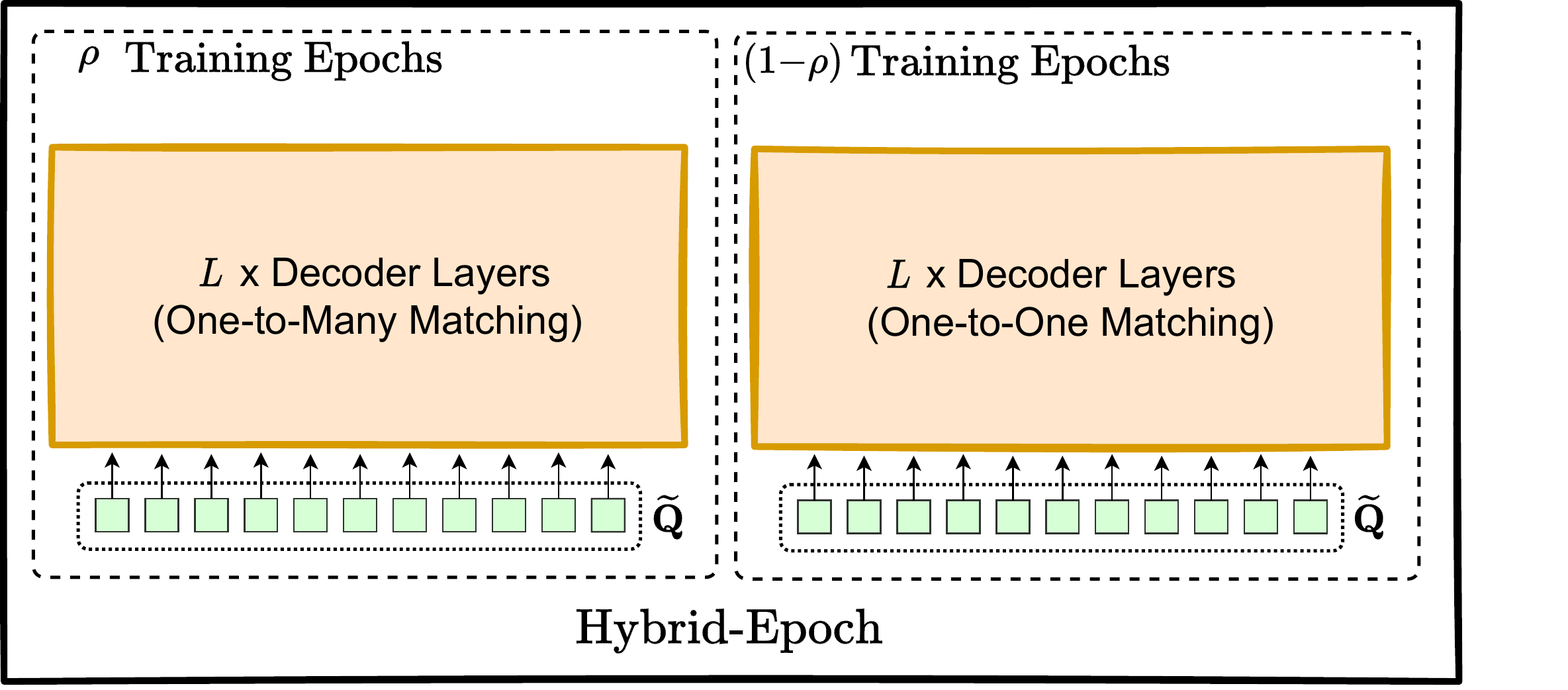}
\end{center}
\end{minipage}
\hspace{-4mm}
\begin{minipage}[t]{0.22\textwidth}
\begin{center}
\includegraphics[width=0.99\textwidth]{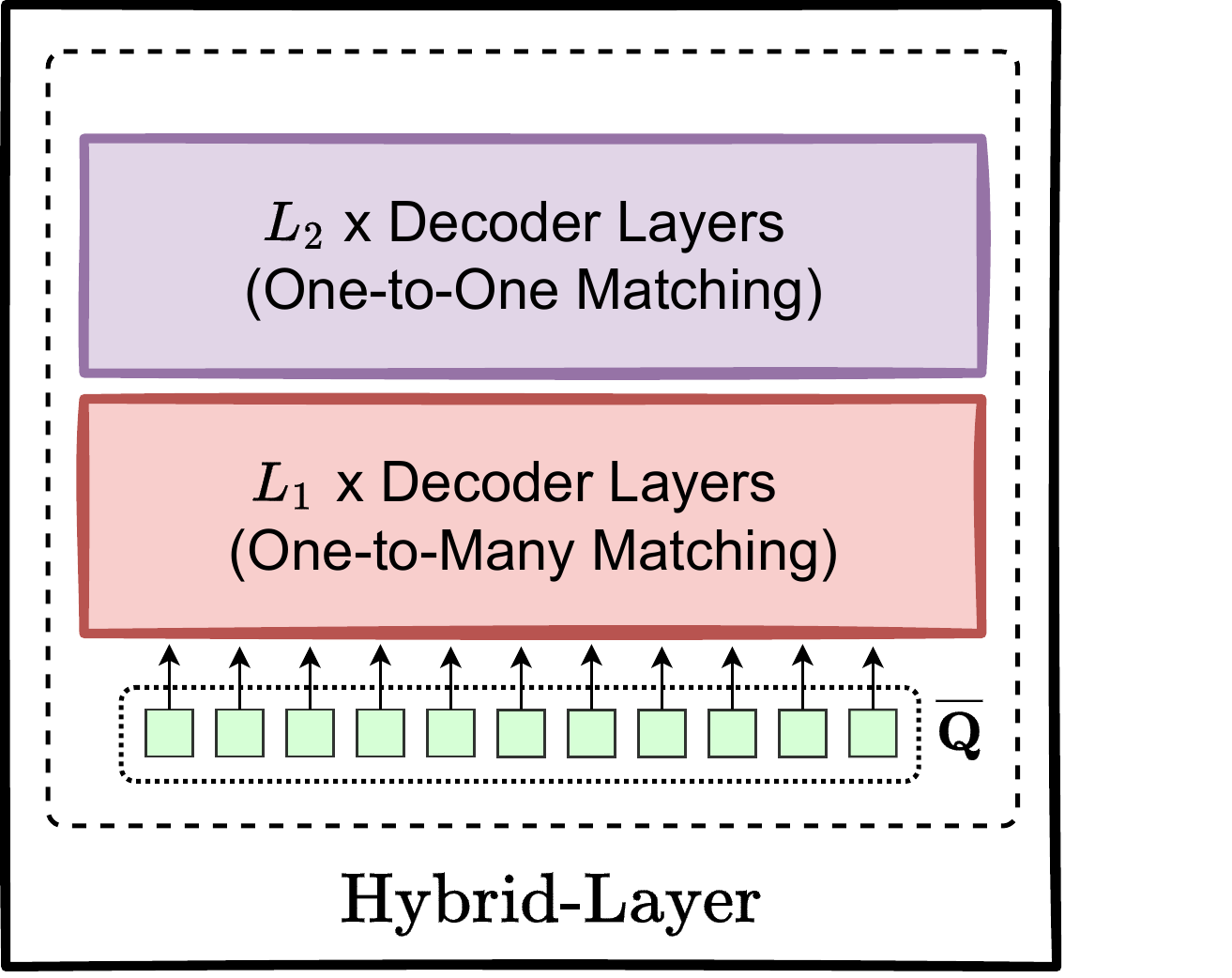}
\end{center}
\end{minipage}
\caption{\small{\textbf{Illustrating the pipeline of our Hybrid Matching scheme.} We use the colored regions of the same color to mark their parameters are shared. We use $\rho$ to represent the percentage of training epochs. We have ${L}{=}{L}_1+{L}_2$ in hybrid layer scheme.}}
\label{fig:hybrid_match_approach}
\vspace{-2mm}
\end{figure*}

The key idea of our hybrid matching approach is to combine the advantages of one-to-one matching
scheme with those of the one-to-many matching scheme,
where the one-to-one matching is necessary for removing NMS and the one-to-many matching
enriches the number of queries that are matched with ground truth for higher training efficacy.
We first illustrate detailed implementations of the hybrid branch scheme, and then briefly introduce the implementations of another two simple variants, including the hybrid epoch scheme and hybrid layer scheme.
We summarize the pipelines of these hybrid matching schemes in Figure~\ref{fig:hybrid_match_approach}.

\vspace{-1mm}
\subsubsection{Hybrid Branch Scheme}

We maintain two groups of queries $\mathbf{Q}{=}\{\mathbf{q}_1,\mathbf{q}_2,\cdots,\mathbf{q}_n\}$ and $\widehat{\mathbf{Q}}{=}\{\widehat{\mathbf{q}}_1,\widehat{\mathbf{q}}_2,\cdots,\widehat{\mathbf{q}}_T\}$, where we apply one-to-one matching or one-to-many matching on the predictions based on $\mathbf{Q}$ or $\widehat{\mathbf{Q}}$ respectively.

\vspace{1mm}
\noindent\textbf{One-to-one matching branch.}
We process the first group of queries $\mathbf{Q}$ with $L$ transformer decoder layers and perform predictions on the output of each decoder layer respectively. Then, we perform the bipartite matching between the \{predictions, ground-truth\} pair over each layer, e.g., estimating $\lmatch{\mathbf{P}^{l}, \mathbf{G}}$, and compute the loss as follows:
\begin{equation}
    \mathcal{L}_{\rm{one2one}} = \sum_{l=1}^{L} \hloss{\mathbf{P}^{l}, \mathbf{G}},\label{eq:loss_one2one_hybrid_branch}
\end{equation}
where $\mathbf{P}^{l}$ represents the predictions outputted by the $l$-th transformer decoder layer.
We choose $\lmatch{\cdot}$ and $\hloss{\cdot}$ following DETR~\cite{carion2020end} and Deformable-DETR~\cite{zhu2020deformable},
which consist of a classification loss, a $\mathcal{L}_1$ regression loss, and a GIoU loss.

\vspace{2mm}
\noindent\textbf{One-to-many matching branch.}
Then, we process the second group of queries $\widehat{\mathbf{Q}}$ with the same $L$ transformer decoder layers and get $L$ groups of predictions.
In order to perform one-to-many matching, we simply repeat the ground truth for $K$ times and get an augmented target $\widehat{\mathbf{G}}{=}\{\mathbf{G}^1,\mathbf{G}^2,\cdots,\mathbf{G}^{K}\}$, where $\mathbf{G}^1{=}\mathbf{G}^2{=}\cdots{=}\mathbf{G}^{K}{=}\mathbf{G}$.
We also perform the bipartite matching between the \{predictions, augmented ground truth\} pair over each layer, e.g., estimating $\lmatch{\widehat{\mathbf{P}}^{l}, \widehat{\mathbf{G}}}$, and compute the corresponding loss as follows:
\begin{equation}
    \mathcal{L}_{\rm{one2many}} = \sum_{l=1}^L \hloss{\widehat{\mathbf{P}}^{l}, \widehat{\mathbf{G}}},\label{eq:loss_one2many_hybrid_branch}
\end{equation}
where $\widehat{\mathbf{P}}^{l}$ represents the predictions output by the $l$-th transformer decoder layer.

In summary, we use the combination of the above two losses, i.e., $\lambda\mathcal{L}_{\rm{one2many}}{+}\mathcal{L}_{\rm{one2one}}$, through the whole training process.
To accelerate the training speed and process both $\mathbf{Q}$ or $\widehat{\mathbf{Q}}$ in parallel, we further apply a masked multi-head self-attention to avoid their interactions.
Therefore, we do not observe significant extra training costs in experiments.
We provide detailed comparison results in the following Table~\ref{tab:compare_hybrid_approaches} within the experiment section.
Last, we only keep the one-to-one matching branch, i.e., ${\mathbf{Q}}$, during evaluation.
We present the overall pipeline on the left of Figure~\ref{fig:hybrid_match_approach}.

\vspace{-3mm}
\subsubsection{More Variants of Hybrid Matching}

\noindent\textbf{Hybrid epoch scheme.}
Different from the hybrid branch scheme, we only maintain a single group of queries $\widetilde{\mathbf{Q}}{=}\{\widetilde{\mathbf{q}}_1,\widetilde{\mathbf{q}}_2,\cdots,\widetilde{\mathbf{q}}_{{M}}\}$, where we apply both one-to-one matching and one-to-many matching on the predictions based on $\widetilde{\mathbf{Q}}$ during different training epochs.
We illustrate more details as follows.

\vspace{1mm}
\noindent\emph{- One-to-many matching training epochs}:
During the first $\rho$ training epochs, we perform one-to-many matching to process $\widetilde{\mathbf{Q}}$ with $L$ transformer decoder layers and get $L$ groups of predictions.
We also get the augmented ground truth $\widetilde{\mathbf{G}}{=}\{\mathbf{G}^1,\mathbf{G}^2,\cdots,\mathbf{G}^{\widetilde{K}}\}$ following the similar manner adopted by the one-to-many matching branch.
Then we perform the bipartite matching between $\lmatch{\widetilde{\mathbf{P}}^{l}, \widetilde{\mathbf{G}}}$ and compute the loss as follows:
\begin{equation}
    \mathcal{L}_{\rm{one2many}} = \sum_{l=1}^{L} \hloss{\widetilde{\mathbf{P}}^{l}, \widetilde{\mathbf{G}}}.\label{eq:loss_one2many_hybrid_epoch}
\end{equation}

\noindent\emph{- One-to-one matching training epochs}:
We change one-to-many matching to one-to-one matching for the remaining $(1-\rho)$ training epochs. The only difference is that we match the predictions with the original ground truth and illustrate the formulation as follows:
\begin{equation}
    \mathcal{L}_{\rm{one2one}} = \sum_{l=1}^{L} \hloss{\widetilde{\mathbf{P}}^{l}, {\mathbf{G}}}.\label{eq:loss_one2one_hybrid_epoch}
\end{equation}

Last, we directly apply $\widehat{\mathbf{Q}}$ during evaluations without using NMS.
In summary, we apply only $\mathcal{L}_{\rm{one2many}}$ or $\mathcal{L}_{\rm{one2one}}$ in the first $\rho$ training epochs or last $(1{-}\rho)$ training epochs respectively.
We also illustrate the overall pipeline of the hybrid epoch scheme in the middle of Figure~\ref{fig:hybrid_match_approach}.

\vspace{2mm}
\noindent\textbf{Hybrid layer scheme.}
Similar to the hybrid epoch scheme, we also maintain a single group of queries $\overline{\mathbf{Q}}{=}\{\overline{\mathbf{q}}_1,\overline{\mathbf{q}}_2,\cdots,\overline{\mathbf{q}}_{N}\}$.
Instead of performing different matching strategies across different training epochs,
we apply one-to-many matching on the prediction output by the first $L_1$ transformer decoder layers and one-to-one matching on the prediction output by the remaining $L_2$ transformer decoder layers.

\vspace{1mm}
\noindent\emph{- One-to-many matching decoder layers}:
We choose to apply a one-to-many matching scheme for the first $L_1$ transformer decoder layers,
where we supervise the predictions, output by each one of the first $L_1$ layers, with the augmented ground truth $\overline{\mathbf{G}}{=}\{\mathbf{G}^1,\mathbf{G}^2,\cdots,\mathbf{G}^{\overline{K}}\}$ following:
\begin{equation}
    \mathcal{L}_{\rm{one2many}} = \sum_{l=1}^{L} \hloss{\overline{\mathbf{P}}^{l}, {\overline{\mathbf{G}}}},\label{eq:loss_one2many_hybrid_layer}
\end{equation}
where we also need to perform the bipartite matching between $\lmatch{\overline{\mathbf{P}}^{l}, \overline{\mathbf{G}}}$ before computing the above loss.

\vspace{1mm}
\noindent\emph{- One-to-one matching decoder layers}:
For the following $L_2$ transformer decoder layers,
we perform a one-to-one matching scheme on their predictions as follows:
\begin{equation}
    \mathcal{L}_{\rm{one2one}} = \sum_{l={{L}_1}}^{{L}_1+{L}_2} \hloss{\overline{\mathbf{P}}^{l}, {\mathbf{G}}}.\label{eq:loss_one2one_hybrid_layer}
\end{equation}

In summary, we apply the combination of both $\mathcal{L}_{\rm{one2many}}$ and $\mathcal{L}_{\rm{one2one}}$ through the whole training procedure.
The right of Figure~\ref{fig:hybrid_match_approach} presents the overall pipeline.

\section{Experiment}

\vspace{1mm}
\subsection{Improving DETR-based Approaches}
\vspace{1mm}

\noindent{\textbf{{2D object detection results.}}}
Table~\ref{tab:improve_2ddet_coco} reports the comparison results on the COCO object detection \texttt{val} set.
Our $\mathcal{H}$-Deformable-DETR achieves consistent gains over the baseline with backbones of different scales (including ResNet-$50$, Swin-T and Swin-L) trained under $12$ epochs or $36$ epochs.
For example, when choosing Swin-T under $12$ and $36$ training epochs, our $\mathcal{H}$-Deformable-DETR improves Deformable-DETR from $51.8\%$ to $53.2\%$ and $49.3\%$ to $50.6\%$, respectively.

Besides, we report the comparison results on LVIS object detection in Table~\ref{tab:compare_to_sota_2ddet_lvis}.
Our approach also achieves consistent gains over the baseline Deformable-DETR across various backbones, e.g., with Swin-L as the backbone, $\mathcal{H}$-Deformable-DETR improves the AP score by $+0.9\%$.

\begin{table}[t]
\begin{minipage}[t]{1\linewidth}
\newcommand{\blue}[1]{\textcolor{blue}{#1}}
\definecolor{deepgreen}{rgb}{0.07, 0.53, 0.03}
\centering
\setlength{\tabcolsep}{2pt}
\renewcommand{\arraystretch}{1.35}
\footnotesize
\caption{\small{Object detection results on COCO.}
\vspace{-1mm}
}
\label{tab:improve_2ddet_coco}
\resizebox{1.0\linewidth}{!}
{
\begin{tabular}{l|c|c|cccc}
\shline
method    & backbone & \#epochs & AP &  AP$_{S}$ & AP$_{M}$ & AP$_{L}$\\
\shline
\multicolumn{7}{l}{\emph{Results under $1\times$ training schedule}}  \\\hline
Deformable-DETR                                                & R$50$   & $12$     & $47.0$ & $29.1$ & $50.0$ & $61.6$ \\
$\mathcal{H}$-Deformable-DETR                & R$50$   & $12$     & $\bf{48.7}^{\textrm{+1.7}}$ & $31.2$ & $51.5$ & $63.5$ \\\hline
Deformable-DETR                                                & Swin-T  & $12$     & $49.3$ &  $31.6$ & $52.4$ & $64.6$\\
$\mathcal{H}$-Deformable-DETR                & Swin-T  & $12$     & $\bf{50.6}^{\textrm{+1.3}}$ & $33.4$ & $53.7$ & $65.9$ \\\hline
Deformable-DETR                                                & Swin-L  & $12$     & $54.5$ &  $37.0$ & $58.6$ & $71.0$ \\
$\mathcal{H}$-Deformable-DETR                & Swin-L  & $12$     & $\bf{55.9}^{\textrm{+1.4}}$ & $39.1$ & $59.9$ & $72.2$  \\\hline
\multicolumn{7}{l}{\emph{Results under $3\times$ training schedule}}                            \\\hline
Deformable-DETR                                                & R$50$   & $36$     & $49.0$ & $32.6$ & $52.3$ & $63.3$ \\
$\mathcal{H}$-Deformable-DETR                & R$50$   & $36$     & $\bf{50.0}^{\textrm{+1.0}}$ & $32.9$ & $52.7$ & $65.3$ \\\hline
Deformable-DETR                                                & Swin-T  & $36$     & $51.8$ & $34.8$ & $55.1$ & $67.8$ \\
$\mathcal{H}$-Deformable-DETR                & Swin-T  & $36$     & $\bf{53.2}^{\textrm{+1.4}}$ & $35.9$ & $56.4$ & $68.2$\\\hline
Deformable-DETR                                                & Swin-L  & $36$     & $56.3$ & $39.2$ & $60.4$ & $71.8$\\
$\mathcal{H}$-Deformable-DETR                & Swin-L  & $36$     & $\bf{57.1}^{\textrm{+0.8}}$ & ${39.7}$ & ${61.4}$ & ${73.4}$\\\shline
\end{tabular}
}
\vspace{3mm}
\end{minipage}
\begin{minipage}[t]{1\linewidth}
    \centering\setlength{\tabcolsep}{2pt}
    \footnotesize
    \renewcommand{\arraystretch}{1.35}
    \caption{\small{Object detection results on LVIS v$1.0$.}
    }
    \resizebox{1.0\linewidth}{!}
    {
        \begin{tabular}{l|c|c|cccc}
            \shline
            method                                             & backbone        & \#epochs & AP  & AP$_{S}$ & AP$_{M}$ & AP$_{L}$               \\
            \shline
            Deformable-DETR                                   & R$50$           & $24$     & $32.2$  & $23.2$ & $41.6$ & $49.3$  \\
            $\mathcal{H}$-Deformable-DETR   & R$50$           & $24$     & $\bf{33.5}^{\textrm{+1.3}}$ & $24.1$ & $42.4$ & $50.2$  \\ \hline
            Deformable-DETR                                   & Swin-L          & $ 48 $   & $47.0$ &  $35.9$ & $57.8$ & $66.9$  \\
            $\mathcal{H}$-Deformable-DETR   & Swin-L          & $48$     & $\bf{47.9}^{\textrm{+0.9}}$  & ${36.3}$ & ${58.6}$ & ${67.9}$  \\ 
            \shline
        \end{tabular}
    }
    \label{tab:compare_to_sota_2ddet_lvis}
\end{minipage}
\end{table}

\begin{table}[t]
    \begin{minipage}[t]{1\linewidth}
        \centering\setlength{\tabcolsep}{4pt}
        \footnotesize
        \renewcommand{\arraystretch}{1.2}
        \caption{\small{Multi-view 3D detection results on nuScenes.}}
        \resizebox{1.0\linewidth}{!}
        {
            \begin{tabular}{l|c|c|cc}
                \shline
                {method}                                 & backbone    & \#epochs & mAP                 & NDS                 \\
                \shline
                PETRv$2$~\cite{liu2022petrv2}            & VoVNet-$99$ & $24$     & $41.04$             & $50.25$             \\
                PETRv$2$ ({Our repro.})             & VoVNet-$99$ & $24$     & $40.41$             & $49.69$             \\
                $\mathcal{H}$-PETRv$2$ & VoVNet-$99$ & $24$     & $\bf{41.93}^{\textrm{+1.52}}$           & ${51.23}$           \\ \hline
                PETRv$2$ ({Our repro.})             & VoVNet-$99$ & $36$     & $41.07$             & $50.68$             \\
                $\mathcal{H}$-PETRv$2$ & VoVNet-$99$ & $36$     & $\bf{42.59}^{\textrm{+1.52}}$ & ${52.38}$ \\
                \shline
            \end{tabular}
        }
        \label{tab:multiview_3d_exp}
        \vspace{2mm}
    \end{minipage}
    \begin{minipage}[t]{1\linewidth}
        \centering\setlength{\tabcolsep}{4pt}
        \footnotesize
        \renewcommand{\arraystretch}{1.2}
        \caption{\small{Multi-person pose estimation results on COCO.}}
        \resizebox{1.0\linewidth}{!}
        {
            \begin{tabular}{l|c|c|ccc}
                \shline
                {method}                             & backbone & \#epochs & AP                 & AP$_{M}$           & AP$_{L}$           \\
                \shline
                PETR~\cite{shi2022end}               & R$50$    & $100$    & $68.8$             & $62.7$             & $77.7$             \\
                PETR~({Our repro.})             & R$50$    & $100$    & $69.3$             & $63.3$             & $78.4$             \\
                $\mathcal{H}$-PETR & R$50$    & $100$    & $\bf{70.9}^{\textrm{+1.6}}$ &  ${64.4}$ & ${80.3}$ \\\hline
                PETR~\cite{shi2022end}               & R$101$   & $100$    & $70.0$             & $63.6$             & $79.4$             \\
                PETR~({Our repro.})             & R$101$   & $100$    & $69.9$             & $63.4$             & $79.4$             \\
                $\mathcal{H}$-PETR & R$101$   & $100$    & $\bf{71.0}^{\textrm{+1.1}}$ &  ${64.7}$ & ${80.2}$ \\\hline
                PETR~\cite{shi2022end}               & Swin-L   & $100$    & $73.1$             & $67.2$             & $81.7$             \\
                PETR~({Our repro.})             & Swin-L   & $100$    & $73.3$             & $67.7$             & $81.6$             \\
                $\mathcal{H}$-PETR & Swin-L   & $100$    & $\bf{74.9}^{\textrm{+1.6}}$ & ${69.3}$ & ${83.3}$ \\
                \shline
            \end{tabular}
        }
        \label{tab:pose_detr_exp}
        \vspace{2mm}
    \end{minipage}
    \begin{minipage}[t]{1\linewidth}
        \centering\setlength{\tabcolsep}{3.5pt}
        \footnotesize
        \renewcommand{\arraystretch}{1.2}
        \caption{\small{Multi-object tracking results on MOT.}
        }
        \resizebox{1.0\linewidth}{!}
        {
            \begin{tabular}{l|c|cccc}
                \shline
                method                                      & \#epochs & MOTA ($\uparrow$)              & IDF$1$ ($\uparrow$)              & FN ($\downarrow$)                 \\
                \shline
                \multicolumn{4}{l}{MOT$17$ \texttt{val}}                                                                              \\
                \shline
                TransTrack~\cite{sun2020transtrack}        & $20$     & $67.1$             & $70.3$             & $15820$             \\
                TransTrack~({Our repro.})             & $20$     & $67.1$             & $68.1$             & $15680$             \\
                $\mathcal{H}$-TransTrack & $20$     & $\bf{68.7}^{\textrm{+1.6}}$ & ${68.3}$ & ${13657}$ \\
                \shline
                \multicolumn{4}{l}{MOT$17$ \texttt{test}}                                                                             \\
                \shline
                TransTrack~\cite{sun2020transtrack}        & $20$     & $74.5$             & $63.9$             & $112137$            \\
                TransTrack~({Our repro.})             & $20$     & $74.6$             & $63.2$             & $111105$            \\
                $\mathcal{H}$-TransTrack & $20$     & $\bf{75.7}^{\textrm{+1.1}}$ & ${64.4}$ & ${91155}$ \\
                \shline
            \end{tabular}
            \vspace{2mm}
        }
        \label{tab:trackformer_exp}
    \end{minipage}
    \vspace{-3mm}
\end{table}

\vspace{2mm}
\noindent{\textbf{{3D object detection results.}}}
We choose the very recent representative DETR-based approach, i.e., PETRv$2$~\cite{liu2022petrv2}, to verify the generalization ability of our approach for 3D detection based on multi-view images.
Table~\ref{tab:multiview_3d_exp} summarizes the detailed comparison results.
We can see that our $\mathcal{H}$-PETRv$2$ significantly improves the NDS scores of baseline PETRv$2$ from $50.68\%$ to $52.38\%$ on nuScenes \texttt{val},
thus showing that our hybrid matching improves the localization accuracy of $3$D object detection predictions.
We observe the GPU memory consumption increases from $7235$M to $11175$M, where $78\%$ of the increased GPU memory locates at the cross-attention.
A preliminary optimization by sequential self/cross attention~\cite{liu2022swin} has reduced the memory consumption from $11175$M to $8733$M while maintaining the performance.

\vspace{2mm}
\noindent{\textbf{{Multi-person pose estimation results.}}}
We extend our hybrid matching strategy to the very recent PETR (Pose Estimation with TRansformers)~\cite{shi2022end} and summarize the detailed results in Table~\ref{tab:pose_detr_exp}.
We can see that our approach achieves consistent gains over the baselines.
For example, with Swin-L as the backbone, our $\mathcal{H}$-PETR improves the AP score of PETR from $73.3\%$ to $74.9\%$ on the COCO \texttt{val} under even $100$ training epochs.

\vspace{2mm}
\noindent{\textbf{{Multi-object tracking results.}}}
We apply our hybrid matching scheme to a powerful multi-object tracking approach, e.g., TransTrack~\cite{sun2020transtrack}.
Table~\ref{tab:trackformer_exp} summarizes the detailed comparison results.
We find the results on MOT$17$ suffer from relatively large variance, thus we report the mean performance with ${\sim}3$ runs.
Accordingly, our approach also shows strong potential on the multi-object tracking tasks, and our $\mathcal{H}$-TransTrack improves the MOTA score of TransTrack from $67.1\%$ to $68.7\%$ on MOT$17$ \texttt{val},
where we also observe that the gains are related to much lower false negative rates (FN).

\vspace{2mm}
\noindent{\textbf{{Panoptic segmentation results.}}}
We also report the detailed COCO panoptic segmentation results in the supplementary to verify
the generalization ability of our approach.

\vspace{2mm}
\subsection{Ablation Study}

\vspace{1mm}
\noindent\textbf{Comparing different hybrid matching schemes.}
We first choose a two-stage Deformable-DETR (with increased FFN dimension) as the baseline and compare the results based on our three different hybrid matching approaches.
To ensure fairness,
we choose their settings as follows:

\vspace{1mm}
\noindent\emph{- Baseline settings}:
We use $300$ or $1800$ queries and apply the conventional one-to-one matching following the original Deformable-DETR~\cite{zhu2020deformable}.
{To ensure more fair comparisons with our hybrid branch scheme, we further report the baseline results under $1.3\times$ and $1.15\times$ training epochs to ensure even more total training time than our hybrid branch scheme.
We empirically find that the increased time mainly comes from two aspects including more queries ($1800$ vs. $300$) and the extra Hungarian matching operation and loss computation of one-to-many matching branch.
By simply merging the matching cost computation and loss computation of one-to-one matching branch and one-to-many matching branch, our hybrid-branch scheme only
bring around +$7\%$ extra training time compared to the baseline with $1800$ queries.
The extra time could be further decreased by implementing the Hungarian matching on GPU instead of CPU, which is not the focus of this work.
}

\begin{table}[t]
    \centering
    \setlength{\tabcolsep}{1pt}
    \renewcommand{\arraystretch}{1.5}
    \footnotesize
    \caption{\small{Comparisons of different hybrid matching approach. We ensure that three different hybrid approaches all (i) introduce $6\times$ more positive samples than the baseline, and (ii) use $1800$ query in total. The upper script $\dag$ marks the methods using $1800$ query during both training and evaluation. The time is averaged over all training epochs as a hybrid epoch scheme consists of different training stages. The FPS is tested on the same V$100$ GPU. The upper script $\natural$ marks the time measured with the optimized implementation and we provide more details in the supplementary.}}
    \label{tab:compare_hybrid_approaches}
    \resizebox{1\linewidth}{!}
    {
        \begin{tabular}{l|c|c|c|c|c|c}
            \shline
            \multirow{2}{*}{method}                 & {\multirow{2}{*}{\parbox[c]{1.2cm}{\makecell[c]{ inference                                                            \\ GFLOPs}}}} & {\multirow{2}{*}{\parbox[c]{1.2cm}{\makecell[c]{ train time                                                            \\ (average)}}}} & {\multirow{2}{*}{\parbox[c]{1.2cm}{\makecell[c]{ inference                                                            \\ FPS}}}} & \multicolumn{3}{c}{\#epochs} \\\cline{5-7}
                                                    &                        &                                                       &                   & 12     & 24     & 36                 \\\shline
            Deformable-DETR                         & $\bf{268}$G     & $\bf{65}$min                                   & $\bf{6.7}$ & $43.7$ & $46.4$ & $46.8$             \\
            Deformable-DETR$^{1.3\times}$                         & $\bf{268}$G     & $\bf{65}$min                                   & $\bf{6.7}$ & $44.6$ & $46.3$ & $46.7$\\
            Deformable-DETR{$^\dag$}                & $282$G                 & $75$min                                               & $6.3$             & $44.1$ & $46.6$ & $47.1$             \\
            Deformable-DETR{$^\dag$}{$^{1.15\times}$}                & $282$G                 & $75$min                                               & $6.3$             &    $44.5$  & $46.7$ & $46.9$\\\hline
            Deformable-DETR + Hybrid-Branch         & $\bf{268}$G     & $80$min$^\natural$                                               & $\bf{6.7}$ & $45.9$ & $47.6$ & $\bf{48.0}$ \\
            Deformable-DETR + Hybrid-Epoch{$^\dag$} & $282$G                 & $95$min                                               & $6.3$             & $45.5$ & $47.0$ & $47.9$             \\
            Deformable-DETR + Hybrid-Layer{$^\dag$} & $282$G                 & $100$min                                              & $6.3$             & $45.6$ & $47.9$ & $\bf{48.0}$ \\
            \shline
        \end{tabular}
    }
\end{table}

\begin{figure}[t]
\begin{minipage}[t]{0.485\textwidth}
\centering
\includegraphics[width=0.9\textwidth]{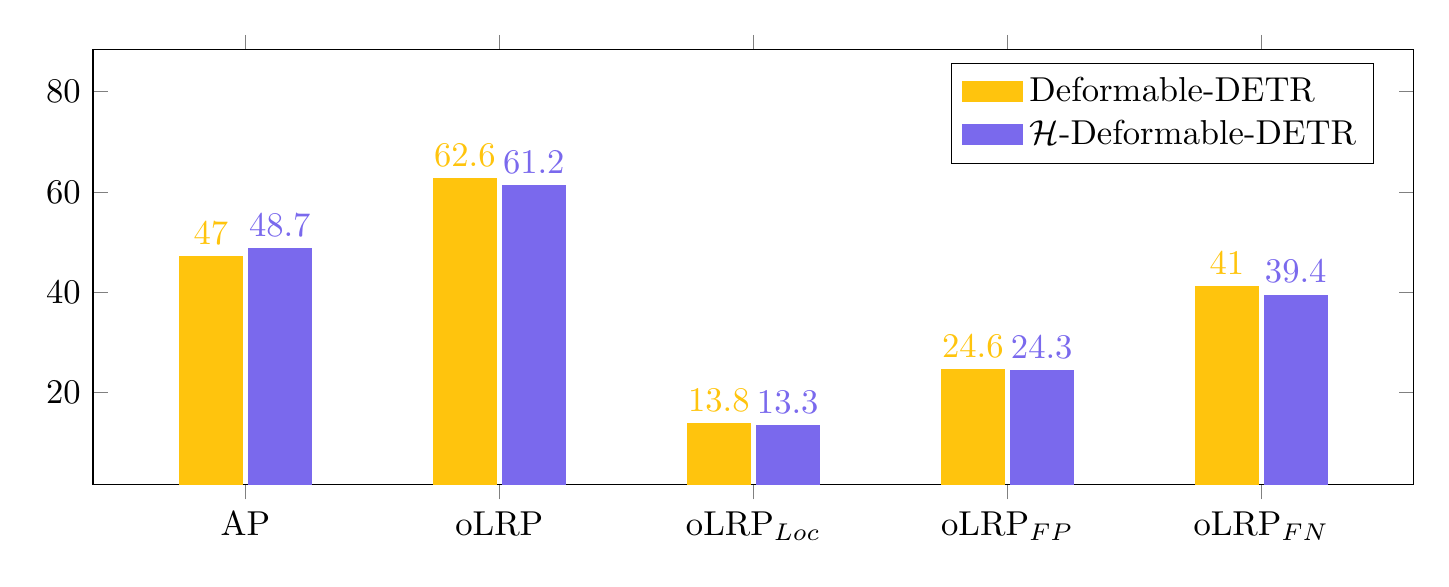}
\vspace{-4mm}
\caption{\small{{Illustrating the improvements of hybrid matching over Deformable-DETR.} We report both average precision (AP) and optimal localization recall precision (oLRP~\cite{LRP_TPAMI}) scores, where the higher the better for AP and the lower the better for oLRP. Our approach significantly decreases {oLRP$_{Loc}$} and {oLRP$_{FN}$}, which represent the localization error of matched detections (true positives) and the number of unmatched ground truth (false negatives).
}}
\vspace{-3mm}
\label{fig:hdetr_improve_olrp}
\end{minipage}
\end{figure}

\vspace{1mm}
\noindent\emph{- Hybrid branch settings}:
We use $300$ queries ($n{=}300$) and $1500$ queries ($T{=}1500$) for one-to-one matching and one-to-many matching branch, respectively.
We set $K$ as $6$ for the augmented ground truth $\widehat{\mathbf{G}}$. Thus we will generate $6\times$ additional positive queries with the one-to-many matching branch, e.g., we have $(1{+}6){\times}12{\times}6$ under $12$ training epochs if the benchmark is the number of positive queries within one training epoch for each decoder layer.

\vspace{1mm}
\noindent\emph{- Hybrid epoch settings}:
We use $1800$ queries through the whole training process, i.e., $M{=}1800$.
To generate $6\times$ additional positive queries than the baseline,
we set ${\widetilde{K}}{=}10$ and $\rho{=}\frac{2}{3}$.
Therefore, we have $10{\times}\frac{2}{3}{\times}12{\times}6{+}\frac{1}{3}{\times}12{\times}6{=}(1{+}6){\times}12{\times}6$ holds under $12$ training epochs.

\vspace{1mm}
\noindent\emph{- Hybrid layer settings}:
We also use $1800$ queries through the whole training process, i.e., $N{=}1800$.
To generate $6\times$ additional positive queries than the baseline,
we set ${\overline{K}}{=}10$, $L_1{=}4$, and $L_2{=}2$ considering that $10{\times}12{\times}4{+}12{\times}2{=}(1{+}6){\times}12{\times}6$ holds.

Table~\ref{tab:compare_hybrid_approaches} summarizes the detailed comparison results.
Accordingly, we find that hybrid matching approaches consistently outperform the baseline under different training epochs (including $1.3\times$ and $1.15\times$ training epochs).
We also report the GPU memory of hybrid-branch/hybrid-layer/hybrid epoch (for ResNet$50$): $7728$M/$7722$M/$7711$M. 
We can see that the hybrid branch scheme is the best choice if we consider the trade-off of training time, inference GFLOPs/FPS, and accuracy.
Therefore, we choose the hybrid branch scheme (for its faster training and inference) by default if not specified in the following ablation experiments.

\begin{table}[t]
\begin{minipage}[t]{1\linewidth}
\centering
\setlength{\tabcolsep}{0.5pt}
\renewcommand{\arraystretch}{1.35}
\footnotesize
\caption{\scriptsize GFLOPs/training time/GPU memory vs. \# query based on our hybrid branch scheme. The numbers within () are based on Swin-L.}
\resizebox{1\linewidth}{!}
{
\begin{tabular}{ l | c | c | c | c | c }
\shline
backbone & method & \# query [hyper-parameter] & GFLOPs & training time (min) & GPU memory (M)   \\ \hline
{\multirow{6}{*}{\parbox[c]{1.1cm}{\makecell{ResNet50                                                \\(Swin-L)}}}}& {\multirow{1}{*}{\parbox[c]{1cm}{\makecell{Baseline}}}}& $300$ [$n$=$300$,$T$=$0$,$K$=$0$] & $268.19$ ($912.29$) & $65$ ($202$) & $5480$ ($8955$)   \\\cline{2-6}
& {\multirow{5}{*}{\parbox[c]{1cm}{\makecell{Ours}}}} & $600$ [$n$=$300$,$T$=$300$,$K$=$6$]  & $271.24$ ($915.34$) & $71$ ($205$) & $5719
$ ($9190$)\\
& & $900$ [$n$=$300$,$T$=$600$,$K$=$6$]  & $274.03$ ($918.12$) &  $72$ ($208$) & $6045$ ($9530$) \\
& & $1200$ [$n$=$300$,$T$=$900$,$K$=$6$]  &$276.82$ ($920.91$) & $75$ ($210$) & $6528$ ($10006$)   \\
& & $1500$ [$n$=$300$,$T$=$1200$,$K$=$6$]  & $279.60$ ($923.69$) & $78$ ($213$) & $7071$ ($10558$)  \\
& & $1800$ [$n$=$300$,$T$=$1500$,$K$=$6$]  & $282.39$ ($926.48$) & $80$ ($215$) & $7728$ ($11203$)   \\
\shline
\end{tabular}}
\label{tab:ablate_gflops_time_gpu}
\end{minipage}
\vspace{-3mm}
\end{table}

\vspace{1mm}
\noindent\textbf{About computation/training time/GPU memory cost.}
Although $\mathcal{H}$-Deformable-DETR uses 6$\times$ number of queries, it has small training overhead on the original Deformable-DETR framework: $1.6\%$/$5.4\%$ on GFLOPs and $6.4\%$/$23.1\%$ on training times for Swin-L/ResNet50, respectively (see Table~\ref{tab:ablate_gflops_time_gpu}). This is because only a portion of the decoder computation is affected by the number of queries, which takes about $0.3\%$/$1.1\%$ of the entire network FLOPs for Swin-L/ResNet$50$. Also note that the more increases in training time is caused by the Hungarian matching (implemented 
on CPU) and the cost/loss computation steps, which are inessential, and we leave its optimization as our future work.

Most of the GPU memory consumption overhead by our method comes from the naive self/cross-attention modules. The large memory complexity is not intrinsic, and it can be significantly reduced by advanced implementation, e.g., Flash Attention~\cite{dao2022flashattention} can decrease the memory to linear complexity, and the sequential attention technique in SwinV2 can also greatly reduce the memory consumption for both self/cross attention. Both implementations are equivalent to the original one, thus not compromising accuracy. With these optimized implementation, the memory consumption increases by more queries are mild.

\vspace{1mm}
\noindent\textbf{Effect of each component based on Deformable-DETR.}
We choose two-stage Deformable-DETR as our baseline and report the detailed improvements of each component in Table~\ref{tab:ablation1}. Notably, we do not observe the obvious benefits of applying drop-out rate as zero, mixed query selection, and looking forward twice\footnote{We implement both mixed query selection and looking forward twice following DINO~\cite{zhang2022dino}.} in other tasks. Therefore, we only apply this combination of tricks on $2$D object detection tasks by default if not specified.
We further analyze the detailed improvements of our approach based on the metrics of optimal localization recall precision in Figure~\ref{fig:hdetr_improve_olrp}, where the lower the better.
In summary,
we can see that our approach mainly improves the performance from two aspects, including more accurate localization and fewer false negatives.

\begin{table}[t]
\begin{minipage}[t]{1\linewidth}
\begin{minipage}[t]{1\linewidth}
\vspace{2mm}
\centering
\setlength{\tabcolsep}{15pt}
\footnotesize
\renewcommand{\arraystretch}{1.2}
\caption{\small{Comparison results based on two-stage Deformable-DETR on COCO 2017 \texttt{val} under $12$ training epochs. $2\times$FFN: increase FFN dimension from $1,024$ to $2,048$. DP$0$: setting the drop out rate within transformer as $0$. MQS: mixed query selection. LFT: look forward twice. HM: our hybrid matching.}
}
\resizebox{1.0\linewidth}{!}
{
    \begin{tabular}{c|c|c|c|c|c}
        \shline
        $2\times$FFN & DP$0$  & MQS    & LFT    & HM     & AP                 \\
        \shline
        \xmark       & \xmark & \xmark & \xmark & \xmark & $43.3$             \\
        \cmark       & \xmark & \xmark & \xmark & \xmark & $43.7$             \\
        \cmark       & \cmark & \xmark & \xmark & \xmark & $44.3$             \\
        \cmark       & \cmark & \cmark & \xmark & \xmark & $46.3$             \\
        \cmark       & \cmark & \cmark & \cmark & \xmark & $47.0$             \\
        \cmark       & \cmark & \cmark & \cmark & \cmark & $\bf{48.7}$ \\
        \shline
    \end{tabular}
}
\label{tab:ablation1}
\end{minipage}

\begin{minipage}[t]{1\linewidth}
\vspace{2mm}
\centering\setlength{\tabcolsep}{6pt}
\footnotesize
\renewcommand{\arraystretch}{1.2}
\caption{\small{Influence of $K$ of our approach on COCO 2017 \texttt{val} under $12$ training epochs. We set $T=300\times K$.}
}
\label{tab:ablation2}
\resizebox{1.0\linewidth}{!}
{
    \begin{tabular}{l|c|c|c|c|c|c|c|c|c}
        \shline
        $K$ & $0$    & $1$    & $2$    & $3$    & $4$    & $5$    & $6$                & $7$    & $8$                \\
        \shline
        AP  & $47.0$ & $46.4$ & $46.7$ & $48.1$ & $48.4$ & $48.3$ & $\bf{48.6}$ & $48.5$ & $\bf{48.6}$ \\
        \shline
    \end{tabular}
}
\end{minipage}
\begin{minipage}[t]{1\linewidth}
\vspace{2mm}
\centering\setlength{\tabcolsep}{12pt}
\footnotesize
\renewcommand{\arraystretch}{1.2}
\caption{\small{Influence of $T$ of our approach on COCO 2017 \texttt{val} under $12$ training epochs. We set $K=6$.}
}
\label{tab:ablation4}
\resizebox{1.0\linewidth}{!}
{
    \begin{tabular}{l|c|c|c|c|c|c}
        \shline
        $T$ & $300$  & $600$  & $900$  & $1200$ & $1500$             & $1800$ \\
        \shline
        AP  & $47.8$ & $48.3$ & $48.4$ & $48.4$ & $\bf{48.7}$ & $48.6$ \\
        \shline
    \end{tabular}
}
\end{minipage}
\end{minipage}
\vspace{-3mm}
\end{table}

\begin{table*}[t]
\begin{minipage}[t]{1\linewidth}
\newcommand{\blue}[1]{\textcolor{blue}{#1}}
    \definecolor{deepgreen}{rgb}{0.07, 0.53, 0.03}
    \centering
    \setlength{\tabcolsep}{7.5pt}
    \renewcommand{\arraystretch}{1.5}
    \footnotesize
    \caption{\small{System-level comparisons with the leading single-scale evaluation results on COCO \texttt{val}.}
        \vspace{-2mm}
    }
    \label{tab:compare_to_sota_2ddet_coco}
    \resizebox{1.0\linewidth}{!}
    {
        \begin{tabular}{l|l|l|l|c|cccccc}
            \shline
            method     & framework & backbone & input size & \#epochs & AP & AP$_{50}$ & AP$_{75}$ & AP$_{S}$ & AP$_{M}$ & AP$_{L}$\\
            \shline
            Swin ~\cite{liu2021swin} & HTC & Swin-L & $1600\times1200$   & $36$ &  {$57.1$} & $75.6$ & $62.5$ & $42.4$ & $60.7$ & $71.1$\\
            CBNetV2~\cite{liang2022cbnet} & HTC & $2\times$ Swin-L & $1600\times1400$  & $12$  & $59.1$ & - & - & - & - & \\
            ConvNeXt~\cite{liu2022convnet} & Cascade Mask R-CNN & ConvNeXt-XL & $1333\times800$   &$36$ & $55.2$ & $74.2$ & $59.9$ & - & - & - \\
            MViTv$2$~\cite{li2021improved} & Cascade Mask R-CNN & MViTv$2$-L & $1333\times800$  & $50$ & $55.8$ & $74.3$ & $64.3$ & - & - & - \\
            MOAT~\cite{yang2022moat} & Cascade Mask R-CNN & MOAT-$3$ & $1344\times1344$  & $36$ & $59.2$ & $\bf77.8$ & $60.9$ & - & - & - \\
            Group-DETR~\cite{chen2022group}  & DETR & Swin-L & $1333\times800$ & $36$ & {$58.4$} & - & - & {$41.0$} & {$62.5$} & {$73.9$} \\
            DINO-DETR~\cite{zhang2022dino}  & DETR & Swin-L & $1333\times800$  & $36$ & {$58.5$} & {$77.0$} & {$64.1$} & {$41.5$} & {$62.3$} & {$74.0$} \\\hline
            $\mathcal{H}$-Deformable-DETR & DETR & Swin-L  & $1333\times800$  & $36$  & {$\bf{59.4}$} & {$\bf{77.8}$} & {$\bf{65.4}$} & {$\bf{43.1}$} & {$\bf{63.1}$} & {$\bf{74.2}$}\\
            \shline
        \end{tabular}
    }
\end{minipage}
\vspace{-2mm}
\end{table*}

\begin{table}[t]
\begin{minipage}[t]{1\linewidth}
\begin{minipage}[t]{1\linewidth}
\centering\setlength{\tabcolsep}{11pt}
\footnotesize
\renewcommand{\arraystretch}{1.2}
\caption{\small{Influence of sharing parameters on COCO 2017 \texttt{val} under $12$ training epochs.}}
\label{tab:ablation10}
\resizebox{1\linewidth}{!}
{
\begin{tabular}{c|c|c|c|c}
    \shline
    trans. encoder & trans. decoder & box head & cls head & AP                 \\
    \shline
    \cmark         & \cmark         & \cmark   & \cmark   & $\bf{48.7}$ \\
    \cmark         & \cmark         & \cmark   & \xmark   & $48.6$             \\
    \cmark         & \cmark         & \xmark   & \xmark   & $48.5$             \\
    \cmark         & \xmark         & \xmark   & \xmark   & $48.3$             \\
    \xmark         & \xmark         & \xmark   & \xmark   & $47.3$             \\
    \shline
\end{tabular}
}
\vspace{3mm}
\end{minipage}
\begin{minipage}[t]{1\linewidth}
\centering
\setlength{\tabcolsep}{3pt}
\renewcommand{\arraystretch}{1.5}
\footnotesize
\caption{\small{Comparison to only using one-to-many matching.}}
\label{tab:compare_to_one2many}
\resizebox{1\linewidth}{!}
{
\begin{tabular}{l|c|c|c|c|c|c}
    \shline
    \multirow{2}{*}{method}   & \multirow{2}{*}{NMS} & {\multirow{2}{*}{\parbox[c]{1.5cm}{\makecell[c]{train time                                                \\(average)}}}} & {\multirow{2}{*}{\parbox[c]{1cm}{\makecell[c]{inference                                                \\FPS}}}} & \multicolumn{3}{c}{\#epochs} \\\cline{5-7}
                              &                      &                                                      &       & 12     & 24                 & 36     \\\shline
    Only one-to-many matching & \cmark               & $95$min                                              & $5.3$ & $\bf{49.4}$ & $\bf{50.2}$ & $48.8$ \\\hline
    Ours (one-to-one branch)  & \xmark               & \multirow{4}{*}{$80$min}                             & $6.7$ & $48.7$ & $49.9$             & $\bf{50.0}$ \\
    Ours (one-to-one branch)  & \cmark               &                                                      & $5.6$ & $48.7$ & $50.0$             & $\bf{50.0}$ \\
    Ours (one-to-many branch) & \xmark               &                                                      & $6.5$ & $13.5$ & $13.1$             & $12.9$ \\
    Ours (one-to-many branch) & \cmark               &                                                      & $5.4$ & $48.6$ & $49.8$             & $49.9$ \\
    \shline
\end{tabular}
}
\end{minipage}
\end{minipage}
\vspace{-3mm}
\end{table}

\vspace{1mm}
\noindent\textbf{Choice of $K$ within the one-to-many matching branch.}
We study the influence of the choice of $K$ in Table~\ref{tab:ablation2}.
Accordingly, we find our approach achieves consistent gains only when choosing $K$ larger than $3$. We choose $K$ as $6$ on COCO by default.
We guess the reasons for the performance drops with smaller $K$ and increases with larger $K$ are: low-quality queries harm accuracy while larger $K$ benefit auxiliary loss which helps accuracy. We verify the quality of different groups of queries between $300$-$1800$ by training a new detector that replaces the original proposals of query $0$-$300$. The APs of queries $300$-$600$,$600$-$900$,$900$-$1200$,$1200$-$1500$,$1500$-$1800$ are $42.7$/$42.8$/$43.0$/$43.1$/$42.5$,
suggesting that different groups have similar qualities, although lower than the default of $0$-$300$ (AP=$47.0$). This indicates the low-quality issue will not get more serious when increasing $K$, while the auxiliary loss can benefit more from larger $K$.
We also show that designing a careful selection mechanism for the one-to-many matching branch can achieve consistent gains even when using $K{=}1$ but do not observe any further benefits when using large $K$ values.
More details are summarized in the supplementary.

\vspace{1mm}
\noindent\textbf{Choice of $T$ within the one-to-many matching branch.}
We study the influence of the total number of queries $T$ within the one-to-many matching branch by fixing $K$ as $6$ in Table~\ref{tab:ablation4}.
We choose $T{=}1500$ on the COCO object detection task as it achieves the best results.

\vspace{1mm}
\noindent\textbf{Effect of sharing parameters.}
We study the influence of the sharing parameters across the one-to-one matching branch and one-to-many matching branch in Table~\ref{tab:ablation10}.
We can observe that (i) using independent classification heads or bounding box heads does not hurt the performance, (ii) further using independent transformer decoder layers slightly drops, and (iii) further using independent transformer encoder layers results in significant performance drops considering the baseline performance is $47.0\%$.
{Accordingly, we can see that {the better optimization of transformer encoder instead of decoder} is the key to the performance improvements.}

{To verify whether this observation still holds on DINO-DETR, we also report the detailed ablation results in the supplementary to verify that most of the gains of (contrast) query denoising
essentially come from {the better optimization of transformer encoder}.
For example, using independent transformer decoders and independent box \& classification heads only suffers from $0.1\%\downarrow$ drop while further using an independent transformer encoder contributes to another $0.5\%\downarrow$ drop.}

\vspace{1mm}
\noindent\textbf{Comparison to only using one-to-many matching.}
Table~\ref{tab:compare_to_one2many} compares our approach to a strong variant that applies only one-to-many matching through the whole training process and NMS during evaluation.
We also ablate the predictions based only on the one-to-one matching branch or one-to-many matching branch in Table~\ref{tab:compare_to_one2many}.
Accordingly, we see that (i) the one-to-many branch evaluated with NMS achieves comparable performance with the one-to-one matching branch; (ii) only using one-to-many matching evaluated with NMS achieves slightly better performance than ours; (ii) the one-to-one matching branch within our hybrid matching scheme is the best choice when considering the training time and inference speed (FPS).

{
We further conduct more ablation experiments including: analyzing the training/validation loss curves of the one-to-one matching branch, the influence of loss-weight on one-to-many matching branch, effect of a careful query selection scheme, precision-recall curves, and so on.
}

\vspace{2mm}
\subsection{Comparison with State-of-the-art}

{
Table~\ref{tab:compare_to_sota_2ddet_coco} reports the system-level comparisons to some representative state-of-the-art methods that use single-scale evaluation on COCO \texttt{val} and choose backbones of similar capacity with Swin-L. By introducing additional enhancement techniques in~\cite{zhang2022dino}, our $\mathcal{H}$-Deformable-DETR achieves $59.4\%$ on COCO \texttt{val} set, which surpasses the very recent DINO-DETR method, as well as other top-performing methods.
}

\section{Conclusion}
This paper presents a very simple yet surprisingly effective hybrid matching scheme to address the low training efficacy of DETR-based approaches on multiple vision tasks.
Our approach explicitly combines the advantages of a one-to-one matching scheme, i.e., avoiding NMS, and those of a one-to-many matching scheme, i.e., increasing the number of positive queries and training efficacy.
We hope our initial efforts can accelerate the advancement of DETR approaches on various vision tasks.

\noindent\textbf{Acknowledgement}
Ding Jia and Chao Zhang are supported by the National Nature Science Foundation of China under Grant 62071013 and 61671027, and National Key R\&D Program of China under Grant 2018AAA0100300.

{\small
\bibliographystyle{ieee_fullname}
\bibliography{egbib}
}

\newpage
\section*{Supplementary}

\subsection*{A. Datasets}

% \vspace{2mm}
\noindent\textbf{COCO}~\cite{lin2014microsoft}. The COCO object detection dataset consists of $123$K images with $896$K annotated bounding boxes belonging to $80$ thing classes and $53$ stuff classes, where the \texttt{train} set contains $118$K images and the \texttt{val} set contains $5$K images. We report the $2$D object detection performance on the \texttt{val} set.
The COCO pose estimation benchmark consists of more than
$200$K images and $250$K person instances labeled with $17$ keypoints,
where \texttt{train} set consists of $57$K images and $150$K person instances, \texttt{val} set consists of $5$K images, and \texttt{test-dev} set consists of $20$K images, respectively.

\vspace{2mm}
\noindent\textbf{LVIS}~\cite{gupta2019lvis}.
This dataset consists of $100$K images annotated with both object detection bounding boxes and instance segmentation masks for $1203$ classes. We follow the very recent work Detic~\cite{zhou2022detecting} to only use the bounding box supervision for training.

\vspace{2mm}
\noindent\textbf{nuScenes}~\cite{caesar2020nuscenes}. This is a large-scale autonomous driving dataset consisting of $1000$ driving sequences with each one about $20$s long, where the multimodel dataset is collected from $6$ cameras, $1$ lidar, and $5$ radars. We partition the $1000$ driving sequences to $700$, $150$, and $150$ sequences for \texttt{train}, \texttt{val}, and \texttt{test} respectively. We report both the numbers of nuScenes Detection Score (NDS) and mean Average Precision (mAP) to measure the $3$D object detection performance.

\vspace{2mm}
\noindent\textbf{ScanNetV$2$}~\cite{dai2017scannet}. This dataset consists of $1513$ indoor scenes labeled with per-point instance ids, semantic categories and $3$D bounding boxes for around $18$ categories. We use $1201$ and $312$ scenes for \texttt{train} and \texttt{val}, respectively. We mainly report the mAP scores under two different IoU thresholds, i.e., $0.25$ and $0.5$.

\vspace{2mm}
\noindent\textbf{MOT17}~\cite{milan2016mot16}. This dataset consists of $14$ pedestrian tracking videos annotated with rich bounding boxes and their corresponding track ids. We use $7$ videos for \texttt{train} and the other $7$ videos for \texttt{test}. Following TransTrack~\cite{sun2020transtrack}, we split the second half of each \texttt{train} video to form a \texttt{val} set. We report the multi-object tracking performance on both \texttt{val} and \texttt{test}.

\vspace{1mm}
\subsection*{B. More Hyper-parameter Details}

We illustrate the detailed hyper-parameter settings when applying our hybrid branch approach to different DETR-based approaches and different benchmarks in Table~\ref{tab:hyper-parameter-detailes}.

\begin{figure*}
\begin{minipage}[t]{1\linewidth}
\begin{center}
\begin{minipage}[t]{0.495\textwidth}
\begin{center}
\includegraphics[width=0.99\textwidth]{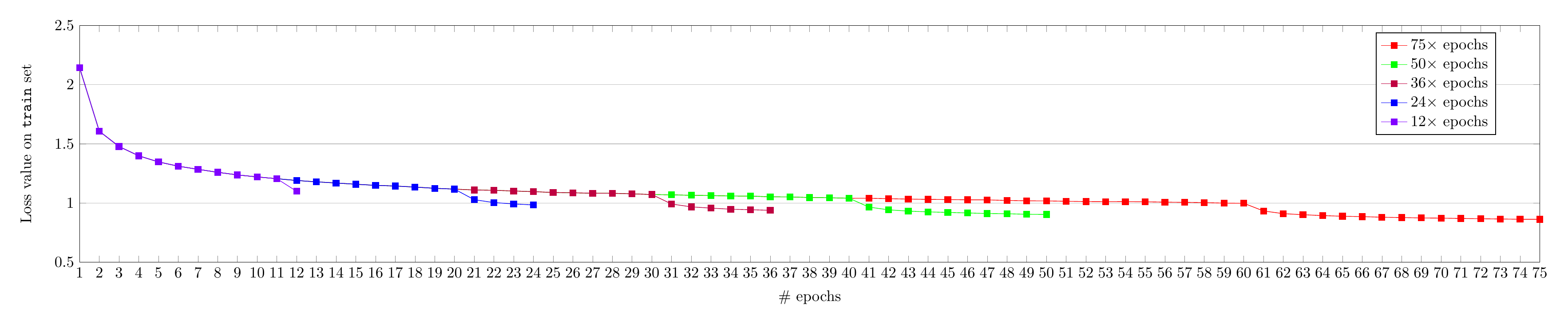}
\end{center}
\end{minipage}
\begin{minipage}[t]{0.495\textwidth}
\begin{center}
\includegraphics[width=0.99\textwidth]{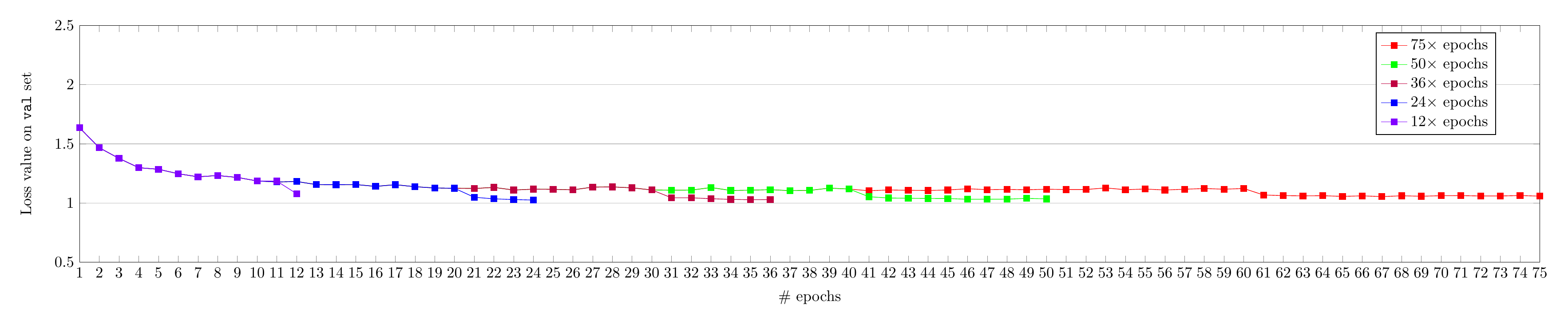}
\end{center}
\end{minipage}
\end{center}
\vspace{-2mm}
\caption{\small{{Illustrating the loss curves of Deformable-DETR on \texttt{train} and \texttt{val}}. We can see that, with longer training epochs, e.g., from $50$ epochs to $75$ epochs, the training loss consistently decreases while the validation loss saturates on COCO object detection benchmark.}}
\label{fig:ddetr_loss_curves}
\end{minipage}
\end{figure*}

\begin{table}[htb]
    \begin{minipage}[t]{1\linewidth}
        \centering
        \footnotesize
        \caption{{Illustrating the hyper-parameter settings. $2\times$FFN: increase FFN dimension from $1,024$ to $2,048$. DP$0$: setting the drop out rate within transformer as $0$. MQS: mixed query selection. LFT: look forward twice.}}
        \setlength\tabcolsep{1pt}
        \renewcommand{\arraystretch}{1.5}
        \resizebox{\linewidth}{!}{
            \begin{tabular}{l|c|c|c|c|c|c|c|c|c}
                \shline
                Method                                         & $2\times$FFN            & DP$0$                   & MQS                     & LFT                     & Dataset  & $n$   & $T$    & $K$ & $\lambda$ \\
                \shline
                \multirow{2}{*}{$\mathcal{H}$-Deformable-DETR} & \multirow{2}{*}{\cmark} & \multirow{2}{*}{\cmark} & \multirow{2}{*}{\cmark} & \multirow{2}{*}{\cmark} & COCO     & $300$ & $1500$ & $6$ & $1.0$     \\\cline{6-10}
                                                               &                         &                         &                         &                         & LVIS     & $300$ & $900$  & $5$ & $1.0$     \\\hline
                \multirow{1}{*}{$\mathcal{H}$-PETR}            & {\cmark}                & {\cmark}                & {\xmark}                & {\xmark}                & COCO     & $300$ & $900$  & $5$ & $1.0$     \\\hline
                $\mathcal{H}$-PETRv$2$                         & {\xmark}                & {\xmark}                & {\xmark}                & {\xmark}                & nuScenes & $900$ & $1800$ & $4$ & $1.0$     \\\hline
                $\mathcal{H}$-TransTrack                       & {\xmark}                & {\xmark}                & {\xmark}                & {\xmark}                & MOT17    & $500$ & $1000$ & $5$ & $0.5$     \\
                \shline
            \end{tabular}
        }
        \label{tab:hyper-parameter-detailes}
        \vspace{-2mm}
    \end{minipage}
\end{table}

\begin{table}[t]
    \centering
    \setlength{\tabcolsep}{10pt}
    \footnotesize
    \renewcommand{\arraystretch}{1.35}
    \caption{\small{Panoptic segmentation results on COCO \texttt{val}.}
    }
    \label{tab:coco_panoptic_exp}
    \resizebox{\linewidth}{!}
    {
        \begin{tabular}{l|c|c|c}
            \shline
            Model                                                & Backbone         & \#epochs & PQ       \\
            \shline
            Mask-Deformable-DETR                                 & R$50$            & $12$     & $47.0$   \\
            $\mathcal{H}$-Mask-Deformable-DETR & R$50$            & $12$     & $\bf{48.5}^{\textrm{+1.5}}$   \\\hline
            Mask-Deformable-DETR                                 & R$50$            & $50$     & $51.5$   \\
            $\mathcal{H}$-Mask-Deformable-DETR & R$50$            & $50$     & $\bf{52.1}^{\textrm{+0.6}}$   \\\hline
            Mask-Deformable-DETR                                 & R$50$            & $100$    & $52.2$   \\
            $\mathcal{H}$-Mask-Deformable-DETR & R$50$            & $100$    & $\bf{52.6}^{\textrm{+0.4}}$ \\\hline
            Mask-Deformable-DETR                                 & Swin-T           & $50$     & $52.9$   \\
            $\mathcal{H}$-Mask-Deformable-DETR & Swin-T           & $50$     & $\bf{53.5}^{\textrm{+0.6}}$ \\\hline
            Mask-Deformable-DETR                                 & Swin-T           & $100$    & $53.8$   \\
            $\mathcal{H}$-Mask-Deformable-DETR & Swin-T           & $100$    & $\bf{54.2}^{\textrm{+0.4}}$ \\\hline
            Mask-Deformable-DETR                                 & Swin-L           & $50$     & $56.7$   \\
            $\mathcal{H}$-Mask-Deformable-DETR & Swin-L           & $50$     & $\bf{57.0}^{\textrm{+0.3}}$ \\\hline
            Mask-Deformable-DETR                                 & Swin-L           & $100$    & $56.9$      \\
            $\mathcal{H}$-Mask-Deformable-DETR & Swin-L           & $100$    & $\bf{57.2}^{\textrm{+0.3}}$     \\
            \shline
        \end{tabular}
    }
\end{table}

\vspace{1mm}
\subsection*{C. Panoptic Segmentation Results}

To verify the effectiveness of our approach to the panoptic segmentation task,
we first construct a simple baseline, i.e., Mask-Deformable-DETR, by adding a mask prediction head to the Deformable-DETR.
Then, we apply our hybrid branch scheme to the Mask-Deformable-DETR following the $\mathcal{H}$-Deformable-DETR, thus resulting in $\mathcal{H}$-Mask-Deformable-DETR.
We summarize the detailed comparison results in Table~\ref{tab:coco_panoptic_exp}.
Accordingly, we can see that our hybrid matching scheme consistently improves ResNet$50$, Swin-T, and Swin-L under various training epochs.

For example, $\mathcal{H}$-Mask-Deformable-DETR gains +$0.3\%$ over Mask-Deformable-DETR equipped with Swin-L when trained for $50$ epochs.We guess the relatively limited gains when compared to object detection tasks, are due to that \emph{the transformer encoder receives much more dense and informative training signals from the extra introduced mask branch}, which consists of the dense interactions between the query embeddings and the high-resolution feature maps output by the transformer encoder. We also have verified that better training of the transformer encoder is the key factor to the performance improvements.

\vspace{1mm}
\subsection*{D. More Ablation Results}

\begin{figure}
\begin{minipage}[t]{0.485\textwidth}
\begin{center}
\includegraphics[width=0.8\textwidth]{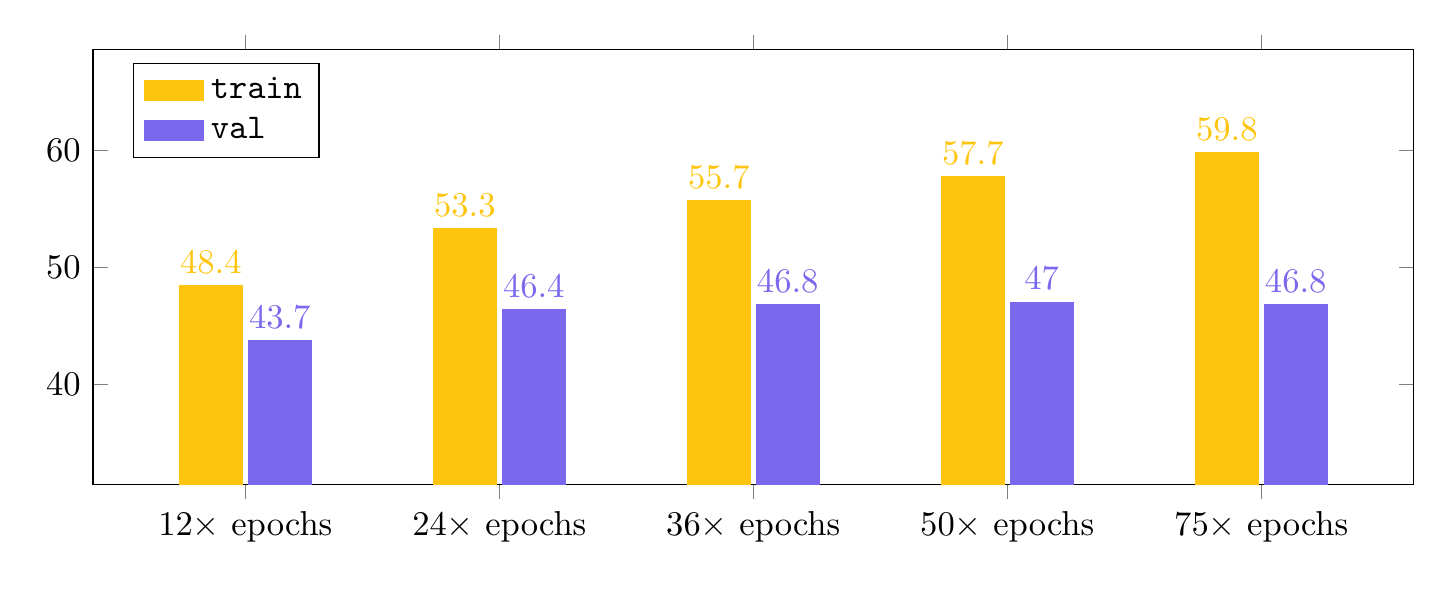}
\end{center}
\vspace{-6mm}
\caption{\small{{Illustrating the AP scores of Deformable-DETR on \texttt{train} and \texttt{val} under longer training epochs}. We can see that, with longer training epochs, e.g., from $50$ epochs to $75$ epochs, the AP scores on \texttt{train} set consistently improves while the AP scores on \texttt{val} set saturates on COCO object detection benchmark.}}
\label{fig:ddetr_ap}
\end{minipage}
\end{figure}

\noindent\emph{-} First, we illustrate the curves of training losses and validation losses based on Deformable-DETR in Figure~\ref{fig:ddetr_loss_curves}.
We also report the detection performance of Deformable-DETR under various training schedules from $12\times$ epochs to $75\times$ epochs in Figure~\ref{fig:ddetr_ap}.
Accordingly, we observe that
(i) simply increasing the number of training epochs fails to improve the performance,
and (ii) Deformable-DETR can not benefit from exploring more positive queries that are matched with ground truth during the additional training epochs, which further verifies the advantage of our hybrid matching scheme.

\begin{table}[t]
    \vspace{2mm}
    \centering\setlength{\tabcolsep}{10pt}
    \footnotesize
    \renewcommand{\arraystretch}{1.2}
    \caption{\small{Influence of $\lambda$ of our approach on COCO 2017 \texttt{val} under $12\times$ epochs training schedule.  We set $K=6$ and $T=1500$.} % 
        \vspace{-2mm}
    }
    \label{tab:ablation3}
    \resizebox{1.0\linewidth}{!}
    {
        \begin{tabular}{l|c|c|c|c|c|c}
            \shline
            $\lambda$ & $0.1$  & $0.2$  & $0.5$  & $1$                & $2$    & $5$    \\
            \shline
            AP        & $47.8$ & $48.0$ & $48.4$ & $\bf{48.7}$ & $48.5$ & $48.3$ \\
            \shline
        \end{tabular}
    }
\end{table}

\begin{table}[t]
\begin{minipage}[t]{1\linewidth}
\centering\setlength{\tabcolsep}{7pt}
\footnotesize
\renewcommand{\arraystretch}{1.35}
\caption{\small{Influence of sharing parameters for DINO framework.\vspace{-2mm}}}
\label{tab:ablate_share_dino}
\resizebox{1\linewidth}{!}
{
\begin{tabular}{c|c|c|c|c}
    \shline
    trans. encoder & trans. decoder & box head & cls head & AP                 \\
    \shline
    \cmark         & \cmark         & \cmark   & \cmark   & ${49.1}$ \\
    \cmark         & \cmark         & \cmark   & \xmark   & $48.9^{\textrm{-0.2}}$             \\
    \cmark         & \cmark         & \xmark   & \xmark   & $49.2^{\textrm{+0.1}}$             \\
    \cmark         & \xmark         & \xmark   & \xmark   & $49.0^{\textrm{-0.1}}$             \\
    \xmark         & \xmark         & \xmark   & \xmark   & $48.5^{\textrm{-0.6}}$             \\
    \shline
\end{tabular}
}
\end{minipage}
\end{table}

\begin{figure}[t]
\begin{minipage}[t]{0.485\textwidth}
    \centering
    \includegraphics[width=0.75\textwidth]{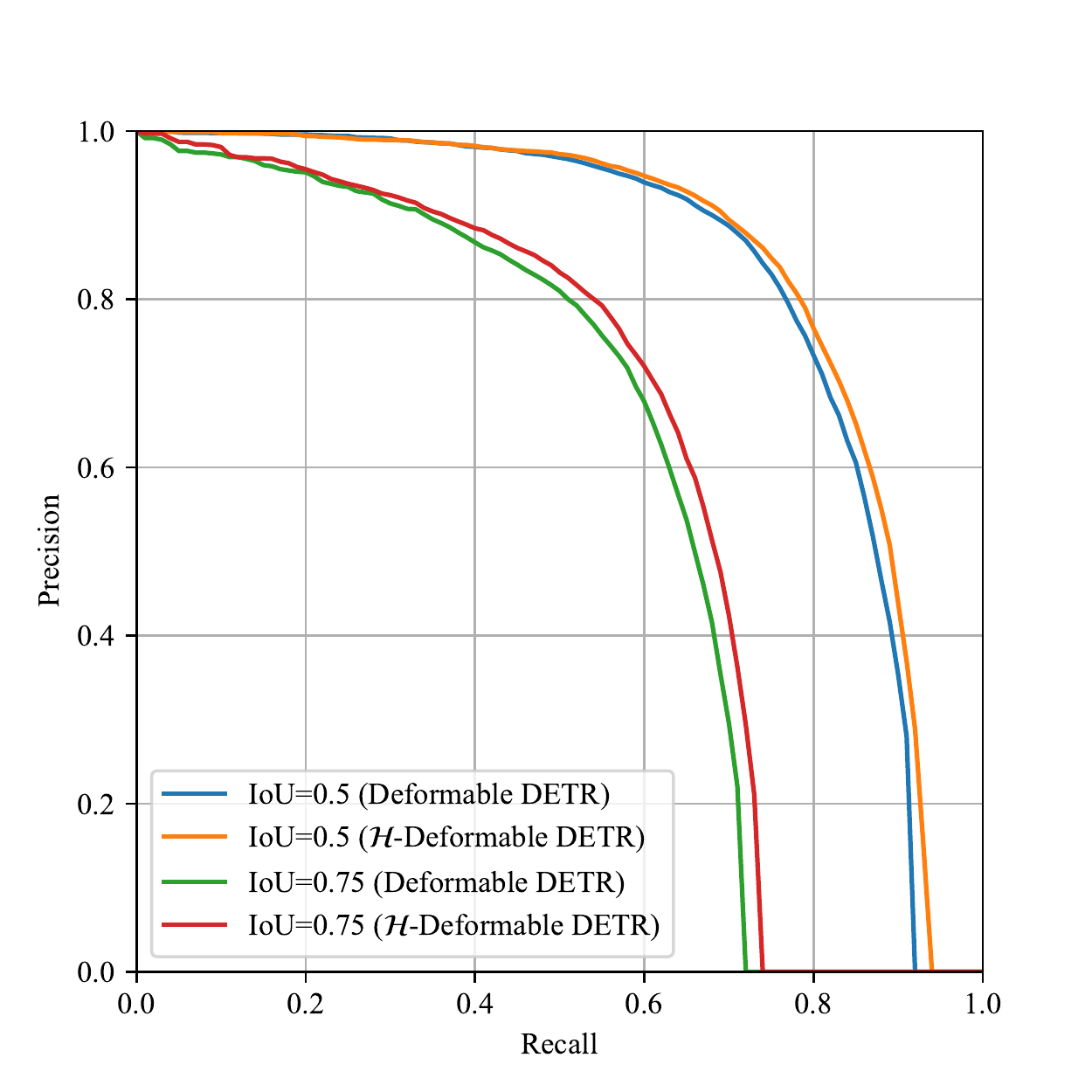}
    \vspace{-2mm}
    \caption{\small{{Illustrating the Precision-Recall curves. We can see that our approach consistently improves the recall of the baseline under two different IoU thresholds}}}
    \label{fig:pr_curves}
\end{minipage}
\end{figure}

\noindent\emph{-} Second, we study the influence of the loss weight $\lambda$ associated with the one-to-many matching loss $\mathcal{L}_{\rm{one2many}}$ in Table~\ref{tab:ablation3}.
We can see that our approach is not sensitive to the choice of $\lambda$ and we simply choose $\lambda{=}1$ for all experiments by default.

\noindent\emph{-} Third, to investigate why hybrid matching fails to improve the performance when choosing $K=1$.
We empirically find that the bounding boxes located at the top $300\sim600$ are of poor localization quality, where simply replacing the top $0\sim300$ bounding box predictions with top $300\sim600$ bounding box predictions during training and evaluation causes a significant performance drop ($47\%$ vs. $42.5\%$).
To address this problem, we study the influence of a more advanced selection scheme within one-to-many matching branch:
(i) use two independent classification heads on the output of the transformer encoder to perform one-to-one/one-to-many (here we use $K+1$ groups of ground truth in one-to-many branch) matching and select top $300$/$300{\times}(K+1)$ predictions respectively,
(ii) use the remaining $300{\times}(K)$ predictions in the one-to-many matching branch by filtering out the duplicated ones presented in the top $300$ predictions of the one-to-one matching branch.
Based on the above-improved selection scheme, we achieve $47.9\%$ when choosing $K=1$ and observe no advantages when choosing $K=6$ compared to the original selection strategy.

\begin{figure*}[t]
\begin{minipage}[t]{1\linewidth}
\begin{center}
\begin{minipage}[t]{0.162\textwidth}
\begin{center}
\includegraphics[width=0.99\textwidth]{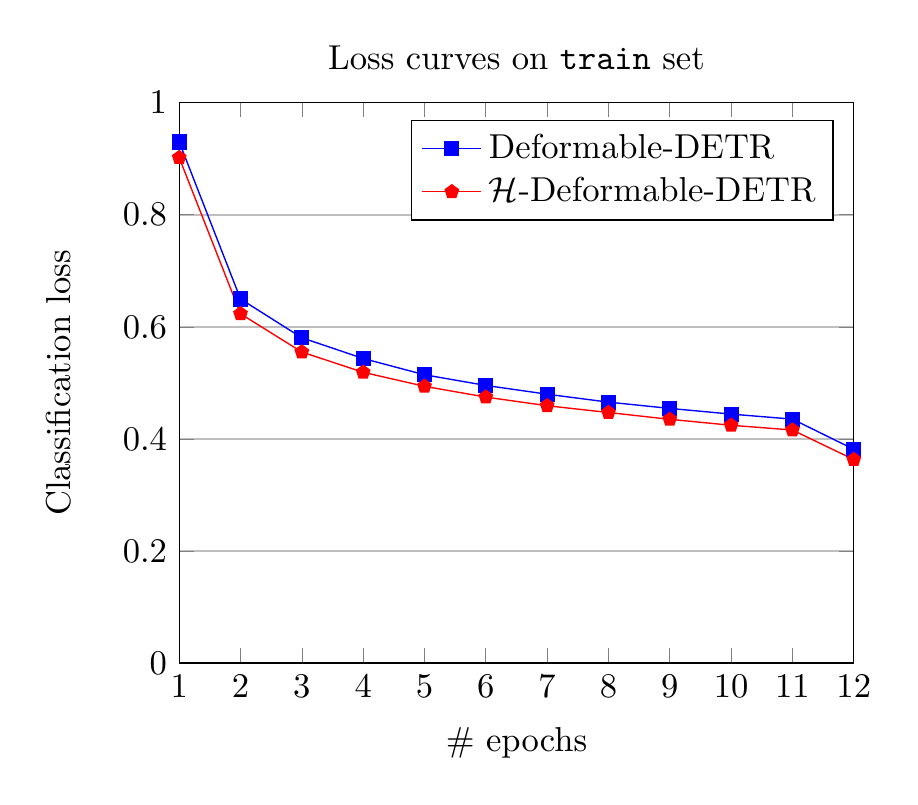}
\end{center}
\end{minipage}
\begin{minipage}[t]{0.162\textwidth}
\begin{center}
\includegraphics[width=0.99\textwidth]{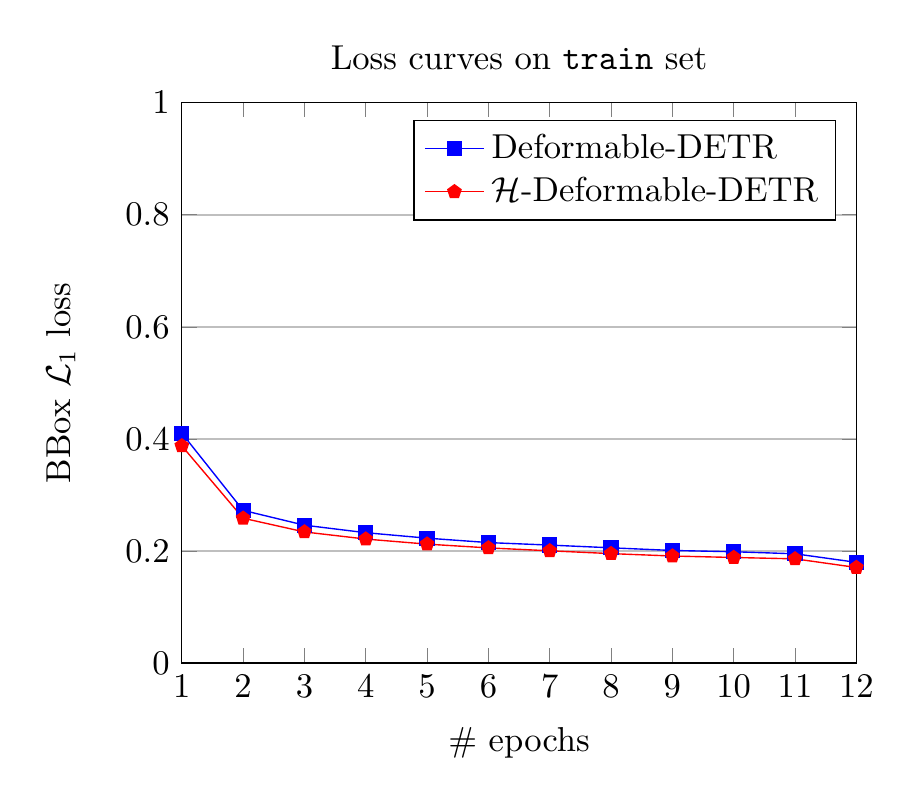}
\end{center}
\end{minipage}
\begin{minipage}[t]{0.162\textwidth}
\begin{center}
\includegraphics[width=0.99\textwidth]{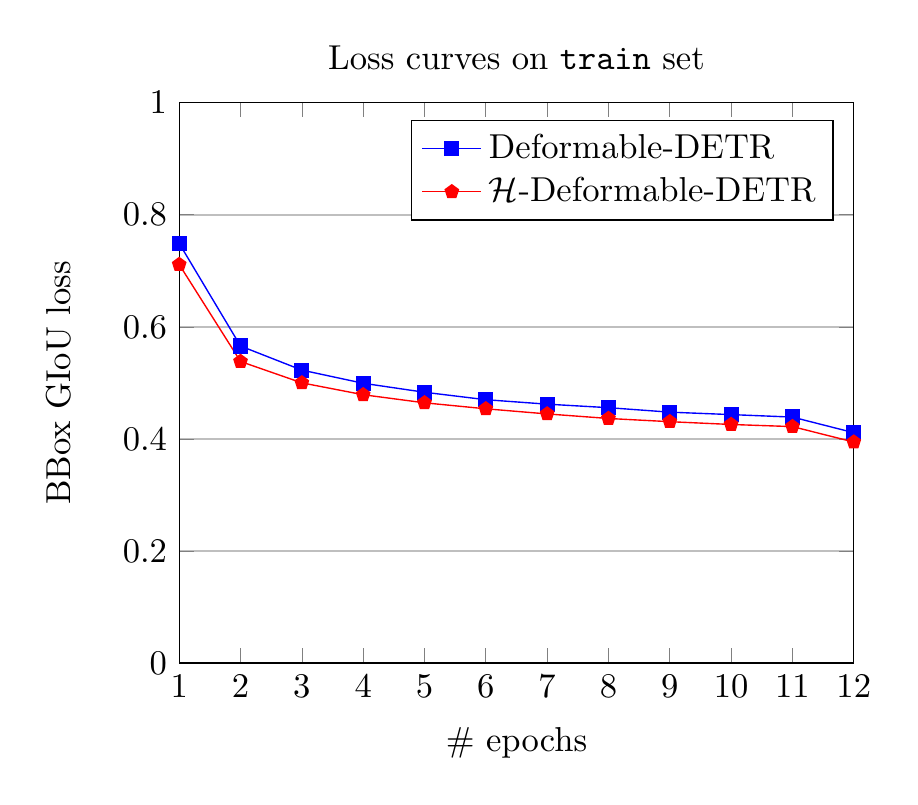}
\end{center}
\end{minipage}
\begin{minipage}[t]{0.162\textwidth}
\begin{center}
\includegraphics[width=0.99\textwidth]{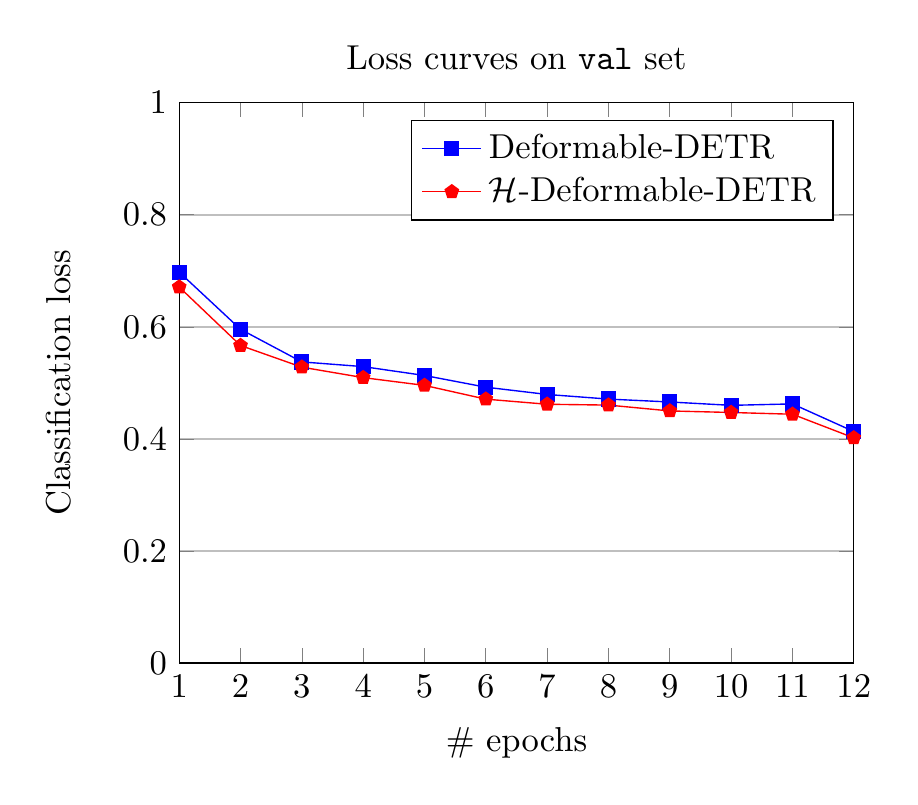}
\end{center}
\end{minipage}
\begin{minipage}[t]{0.162\textwidth}
\begin{center}
\includegraphics[width=0.99\textwidth]{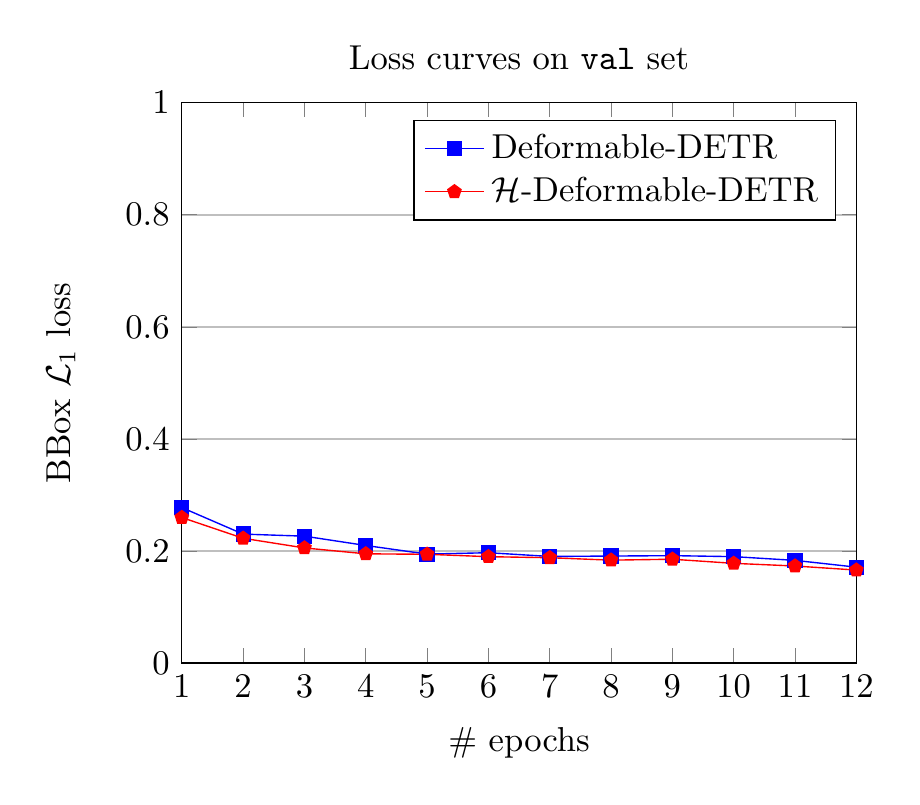}
\end{center}
\end{minipage}
\begin{minipage}[t]{0.162\textwidth}
\begin{center}
\includegraphics[width=0.99\textwidth]{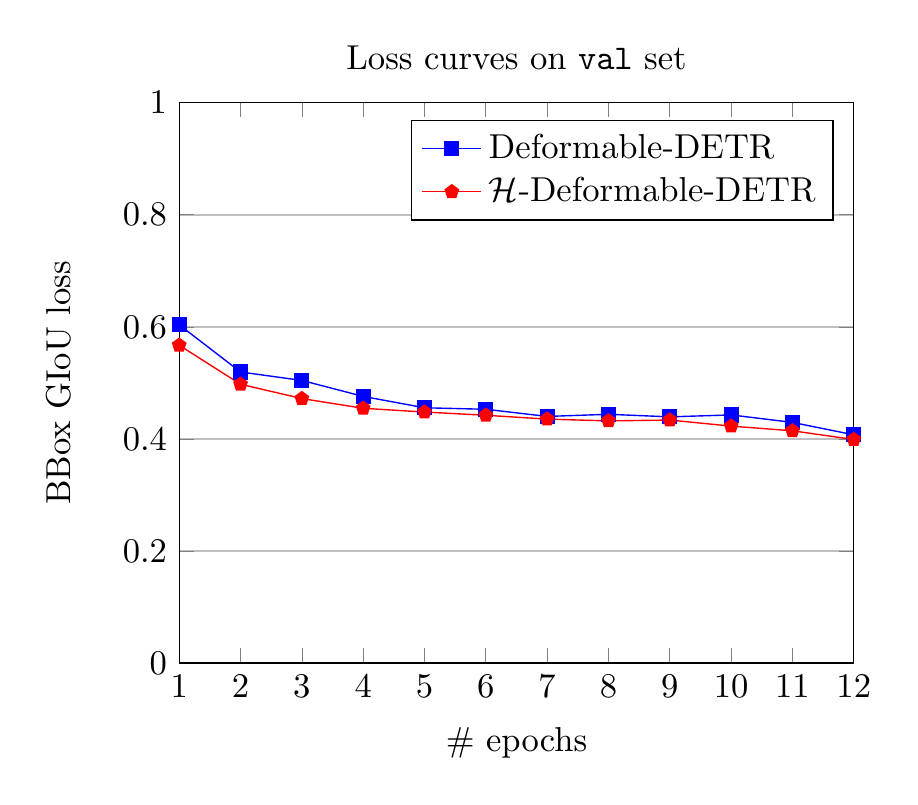}
\end{center}
\end{minipage}
\end{center}
\vspace{-5mm}
\caption{\small{{Illustrating the loss curves on \texttt{train} and \texttt{val}}. Our approach helps the optimization of Deformable-DETR, thus decreasing the loss values on both \texttt{train} and \texttt{val} of COCO object detection benchmark.}}
\label{fig:ablation_loss_curves}
\end{minipage}
\end{figure*}

{
\definecolor{codegreen}{rgb}{0,0.5,0}
\definecolor{codeblue}{rgb}{0.25,0.5,0.5}
\definecolor{codegray}{rgb}{0.6,0.6,0.6}
\lstset{
backgroundcolor=\color{white},
basicstyle=\fontsize{7.5pt}{8.5pt}\fontfamily{lmtt}\selectfont,
columns=fullflexible,
captionpos=b,
commentstyle=\fontsize{8pt}{9pt}\color{codegray},
keywordstyle=\fontsize{8pt}{9pt}\color{codegreen},
stringstyle=\fontsize{8pt}{9pt}\color{codeblue},
frame=none,
otherkeywords = {self},
breaklines=true,
numbers=left,
stepnumber=1,    
firstnumber=1,
numberfirstline=true
}

\begin{algorithm}
\tiny
\begin{lstlisting}[language=python]
def naive_hybrid_loss(
        outputs_one2one,
        outputs_one2many,
        targets,
        k_repeated_targets):
    loss_dict_one2one = criterion(
        outputs_one2one, targets)
    loss_dict_one2many = criterion(
        outputs_one2many, k_repeated_targets)
    def criterion(
        prediction,
        target):
        cost_matrix = compute_cost()
        indices = hungarian_matching(cost_matrix)
        loss(prediction, target, indices)


def optimized_hybrid_loss(
        outputs_one2one,
        outputs_one2many,
        targets,
        k_repeated_targets):
    cost_matrix_one2one, cost_matrix_one2many =
    compute_cost(
        concatenate(outputs_one2one,  outputs_one2many),
        concatenate(targets, k_repeated_targets)
    )
    indices_one2one = hungarian_matching(
        cost_matrix_one2one)
    indices_one2many = hungarian_matching(
        cost_matrix_one2many)
    loss(
        concatenate(outputs_one2one,  outputs_one2many),
        concatenate(target, k_repeated_targets),
        concatenate(indices_one2one, indices_one2many))
\end{lstlisting}
\caption{\small{Hybrid Matching Loss Function.}}
\label{fig:code}
\end{algorithm}
}

\noindent\emph{-} Fourth, to verify whether this observation still holds on DINO-DETR, Table~\ref{tab:ablate_share_dino} reports the detailed ablation results on DINO-DETR.
We can see that the gains of (contrast) query denoising also
mainly comes from the better optimization of the transformer encoder instead of the decoder.
For example, using independent transformer decoders and independent box \& classification heads only suffers from $0.1\%\downarrow$ drop ($49.1\%\to49.0\%$) while further using an independent transformer encoder contributes to another $0.5\%\downarrow$ drop ($49.1\%\to48.5\%$).

\noindent\emph{-} Fifth,
we illustrate the loss curves of our $\mathcal{H}$-Deformable-DETR and the baseline Deformable-DETR in Figure~\ref{fig:ablation_loss_curves}.
To ensure fairness, we only consider the $\mathcal{L}_{\rm{one2one}}$ through the whole training procedure.
Accordingly, we can see our approach achieves both lower training loss values and lower validation loss values,
which shows that the additional one-to-many matching branch could ease the optimization of the one-to-one matching branch.

\noindent\emph{-} Sixth,
we compare the precision-recall curves of the baseline and our hybrid matching scheme in Figure~\ref{fig:pr_curves}.
It can be seen that our approach mainly improves the recall rate of the baseline method, thus also ensuring the lower false negative rates.

\vspace{1mm}
\subsection*{E. Accelerate Hybrid Matching}

In the original implementation, we simply perform the one-to-one/one-to-many matching and loss computation with two independent functions, thus increasing the overall training time of each epoch from $75$min to $85$min when compared
to the baseline that uses the same total number of queries.
To decrease the additional latency, we merge the computation of their cost matrices and loss functions in Algorithm~\ref{fig:code}.
Based on the accelerated implementation, we decrease the original training time from $85$min to $80$min.

\vspace{1mm}
\subsection*{Acknowledgement}
We thank Di He, Zheng Zhang, Jin-Ge Yao, Yue Cao, and Yili Wang for helpful discussions.
Especially, Yili Wang helps to verify the proposed method for multi-person pose estimation tasks at the early stage.

\end{document}